\documentclass[preprint,12pt]{elsarticle}

\usepackage{amssymb}
\usepackage{amsmath}
\usepackage{subcaption}
\usepackage{float} 

\def\sq{^2}
\def\rot{{\sf R}}

\journal{Applications in Engineering Science}

\begin{document}

\begin{frontmatter}



\title{Open-Source Software Architecture for Multi-Robot Wire Arc Additive Manufacturing (WAAM)}


\author[ecse]{Honglu He} 
\author[ecse]{Chen-lung Lu} 
\author[mane]{Jinhan Ren} 
\author[mane]{Joni Dhar} 
\author[mic]{Glenn Saunders} 
\author[wason]{John Wason} 
\author[mane]{Johnson Samuel} 
\author[ecse]{Agung Julius} 
\author[ecse]{John T. Wen} 
\affiliation[ecse]{organization={Electrical, Computer, and Systems Engineering, Rensselaer Polytechnic Institute},
            addressline={110 8th St}, 
            city={Troy},
            postcode={12180}, 
            state={NY},
            country={US}}
\affiliation[mic]{organization={Manufacturing Innovations Center,  Rensselaer Polytechnic Institute},
            addressline={110 8th St}, 
            city={Troy},
            postcode={12180}, 
            state={NY},
            country={US}}
\affiliation[mane]{organization={Mechanical, Aerospace, and Nuclear Engineering, Rensselaer Polytechnic Institute},
            addressline={110 8th St}, 
            city={Troy},
            postcode={12180}, 
            state={NY},
            country={US}}
\affiliation[wason]{organization={Wason Technology, LLC},
            addressline={110 8th St}, 
            city={Troy},
            postcode={12180}, 
            state={NY},
            country={US}}

\begin{abstract}

Wire Arc Additive Manufacturing (WAAM) is a metal 3D printing technology that deposits molten metal wire on a substrate to form desired geometries.  Articulated robot arms are commonly used in WAAM to produce complex geometric shapes.  However, they mostly rely on proprietary robot and weld control software that limits process tuning and customization, incorporation of third-party sensors, implementation on robots and weld controllers from multiple vendors, and customizable user programming.  This paper presents a general open-source software architecture for WAAM that addresses these limitations.  The foundation of this architecture is Robot Raconteur, an open-source control and communication framework that serves as the middleware for integrating robots and sensors from different vendors. 
Based on this architecture, we developed an end-to-end robotic WAAM implementation that takes a CAD file to a printed WAAM part and evaluates the accuracy of the result.  The major components in the architecture include part slicing, robot motion planning, part metrology, in-process sensing, and process tuning.  The current implementation is based on Motoman robots and Fronius weld controller, but the approach is applicable to other industrial robots and weld controllers.  The capability of the WAAM tested is demonstrated through the printing of parts of various geometries and acquisition of in-process sensor data for motion adjustment.
\end{abstract}




\end{frontmatter}



\section{INTRODUCTION}

\begin{figure}[h!]
    \centering
    \includegraphics[width=0.8\textwidth]{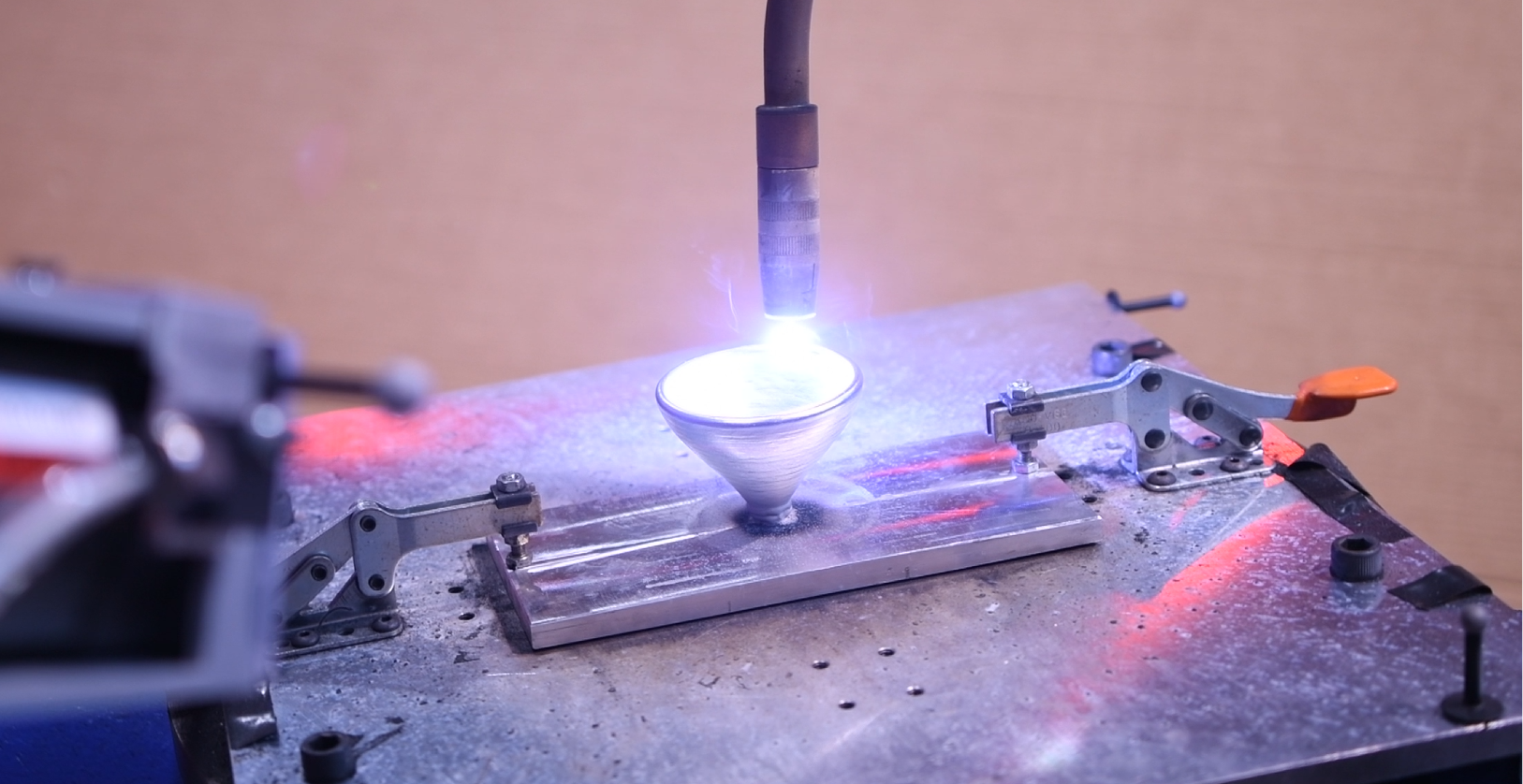}
    \caption{3D Printing using the WAAM Process}
    \label{fig:waam}
\end{figure}

Additive manufacturing has emerged in the past few decades as a powerful tool to manufacture parts of complex geometries \cite{additive_overview}.  Polymer-based 3D printing is now commonplace and offers a convenient user interface and high geometric fidelity. Metal additive manufacturing faces more challenges due to the longer solidification times and different solidification processes for different metal alloys.  Multiple technologies have been developed for metal 3D printing, including powder bed fusion \cite{pbf1}, direct metal deposition \cite{dmd}, cold spray \cite{cold_spray}, and wire arc additive manufacturing (WAAM) \cite{waam_review}.    WAAM adopts the wire arc welding process to deposit molten metal wire feedstock as shown in Fig.~\ref{fig:waam}.  It typically deploys a 6-dof industrial robot with a wrist-mounted weld gun together with a 2-dof positioner to control the flow of the molten metal.  Commercial robotic WAAM systems provide coordinated robot motion and welding arc excitation to achieve the desired metal deposition. Such systems offer a few tunable parameters, such as robot path speed and wire feed rate (usually set to constants), but do not provide convenient tools to integrate third-party process sensors, such as IR cameras, metrology scanners, audio inputs, etc.

There has been numerous studies of various aspects of the WAAM process.
Based on the WAAM bead geometry, research group \cite{bead_geometry} studies on choosing welding parameters for width and height control.  Wider multi-bead deposition is achieved 
through overlapping beads \cite{multi_bead_model} 
or weaving motion deposition \cite{weaving}. Complex geometry WAAM is conducted 
through multi-directional deposition \cite{multidirection_waam4}. 
A WAAM system with CNC welder and 2-dof positioner can produce complex geometry \cite{cnc_waam}. For WAAM geometry evaluation,  ultrasonic sensing is used to examine the part geometry 
\cite{ultrasonic} 
and optical microscope is used to study microstructure properties  \cite{waam_evaluation}.  These studies focus on specific aspects of the WAAM process rather than a general pipeline for the WAAM printing of complex geometric shapes with integrated process monitoring and product evaluation. 

This paper presents an open-source WAAM control architecture with integrated planning, motion control, and sensing that may be adapted to different robot and sensor technologies.  The foundation of the architecture is Robot Raconteur (RR) \cite{RR1}, an open-source communication and control framework for robotics and automation developed by Wason Technology.  Based on this architecture, we developed an art-to-part WAAM implementation, from the CAD model to the final printed part.  The key modules in this implementation include part slicing, motion planning, sensor feedback, and part evaluation.  Process data from sensors and robot motion are acquired at regular rates and may be used for in-process adjustments or post-processing analysis.  Our testbed consists of multiple Motoman robots, a Fronius weld controller, and a suite of sensors.  The robots include a 6-dof welding robot, a two-axis positioner, and a 6-dof process monitoring robot with mounted sensors.  The sensor suite includes FLIR camera, laser scanner, and microphone. 
A 3D scanner provides the metrology of the printed part. RR drivers have been developed for the robots, weld controller, and sensors to enable this integrated system.  We have printed parts of varying geometric complexity and compared the final parts with the corresponding CAD models.  The results show a submillimeter average error. The 
worst-case shape errors for a model fan blade and a bell are under 3~mm and part width variation less than 15\%.  This work contributes to the state of the art of WAAM technology in the development of a
a general software architecture that integrates robots, the weld controller, and process sensors from multiple vendors.  The source code of the complete implementation is available at our repository \cite{welding_motoman}.

We will present an overview of the WAAM process and the overall workflow in Section~\ref{sec:waam_overview}, describe our testbed and software architecture in Section~\ref{sec:architecture}, detail the algorithm design in Section~\ref{sec:algorithm}, and discuss results and assessment in Section~\ref{sec:result}.

\section{WAAM Process}
\label{sec:waam_overview}




\subsection{Process Overview}
As in conventional 3D printing, WAAM uses a spool of metal wire of a given diameter as the primary material feedstock.  An electric arc, as used in traditional welding processes, serves as the heat source to melt the wire feedstock for deposition. As the wire melts, the molten metal is deposited onto a substrate to form the part layer by layer. The deposition process is controlled by the amount of the wire feed rate, robot (and therefore the torch) motion, and the flow of the molten metal under gravity.   The process is guided by the CAD model to achieve a 3D replication of the intended geometry. 
Once the part has been fully created, post-processing may be necessary, especially if the finished product requires enhanced tolerances, surface finishes, or specific mechanical attributes \cite{waam_milling}. Non-destructive testing methods, such as ultrasonic inspections or X-ray imaging \cite{x_ray_inspection}, are frequently employed to verify the integrity and consistency of the finished product.

\subsection{Process Workflow}

\begin{figure}[h]
    \centering
    \includegraphics[width=0.95\textwidth]{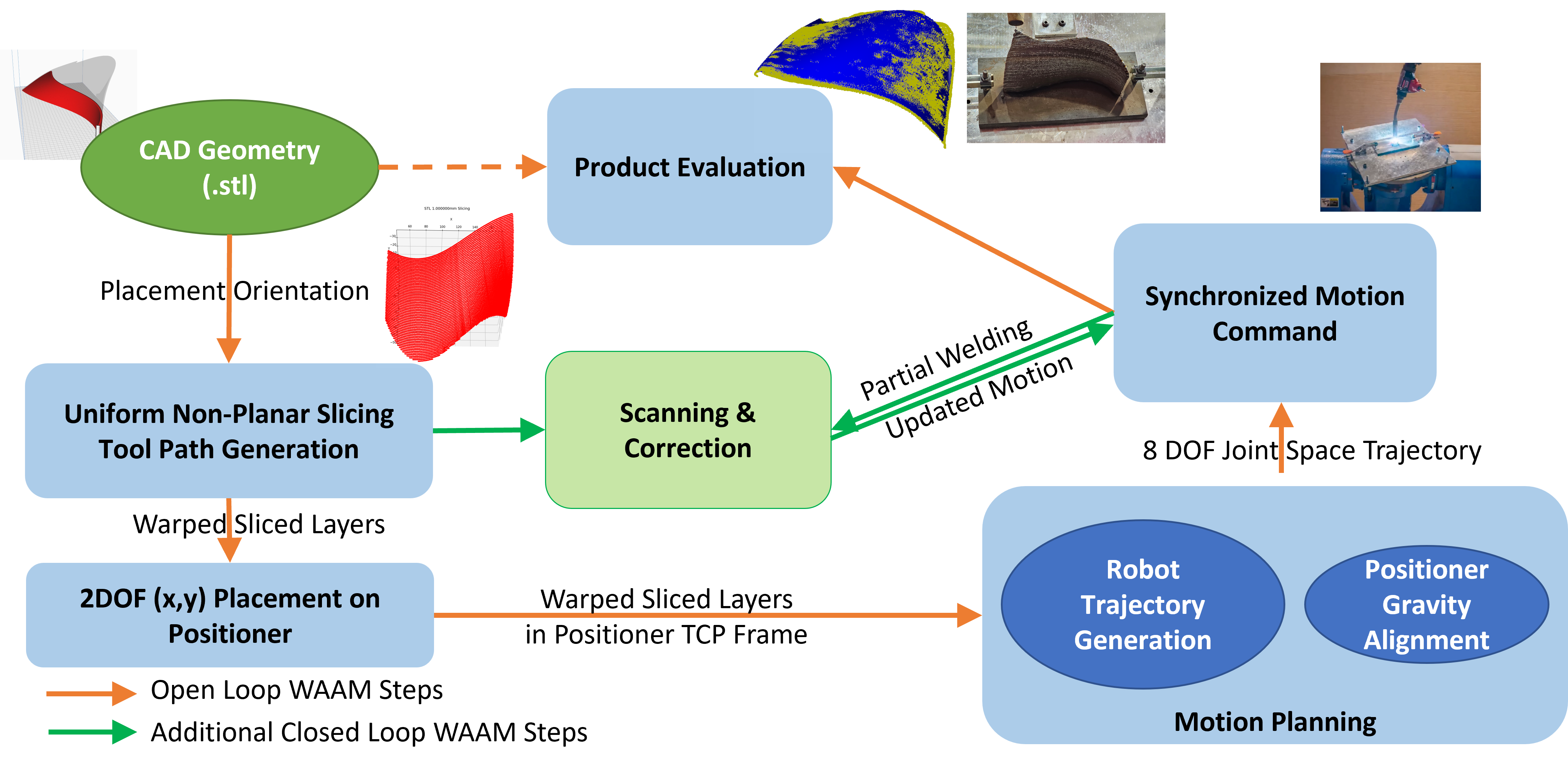}
    \caption{Workflow for Robotic WAAM Process.
    }
    \label{fig:flowchart}

\end{figure}

While WAAM leverages the flexibility of wire arc welding to provide fast prototyping of metal parts, the process poses certain challenges, including non-planar slicing for complex geometries, deposition along gravity, motion planning and coordination of multiple robots, and selection of weld controller parameters. 
Fig.~\ref{fig:flowchart} shows the key steps in an end-to-end pipeline of the WAAM process.  
\begin{itemize}
    \item {\em Slicing}: Decompose the CAD geometry to slices of uniform height suitable for layer-by-layer printing.
    \item {\em Part Placement}: Choose the position and orientation of the desired part on the 2-dof positioner relative to the 6-dof welding robot.
    \item {\em Robot Motion Planning}: Coordinate the motion of the welding robot and the positioner motion by specifying motion waypoints and path speed.  The deposition direction should always point towards the downward gravitational acceleration to avoid uncontrolled flowing of the molten metal.
    \item {\em In-Process Sensor Monitoring}: Collect in-process sensor data to guide process tuning and robot motion adjustment.
    \item {\em Post-Printing Metrology}: Perform 3D scan of the printed part and compare with the desired CAD model.
\end{itemize}
To set up the overall WAAM cell, it is also important to perform robot calibration (robot kinematics, and the relative pose between welding robot and positioner).  


%


%

\section{System Architecture}
\label{sec:architecture}

\begin{figure*}
    \centering
    \includegraphics[width=0.9\textwidth]{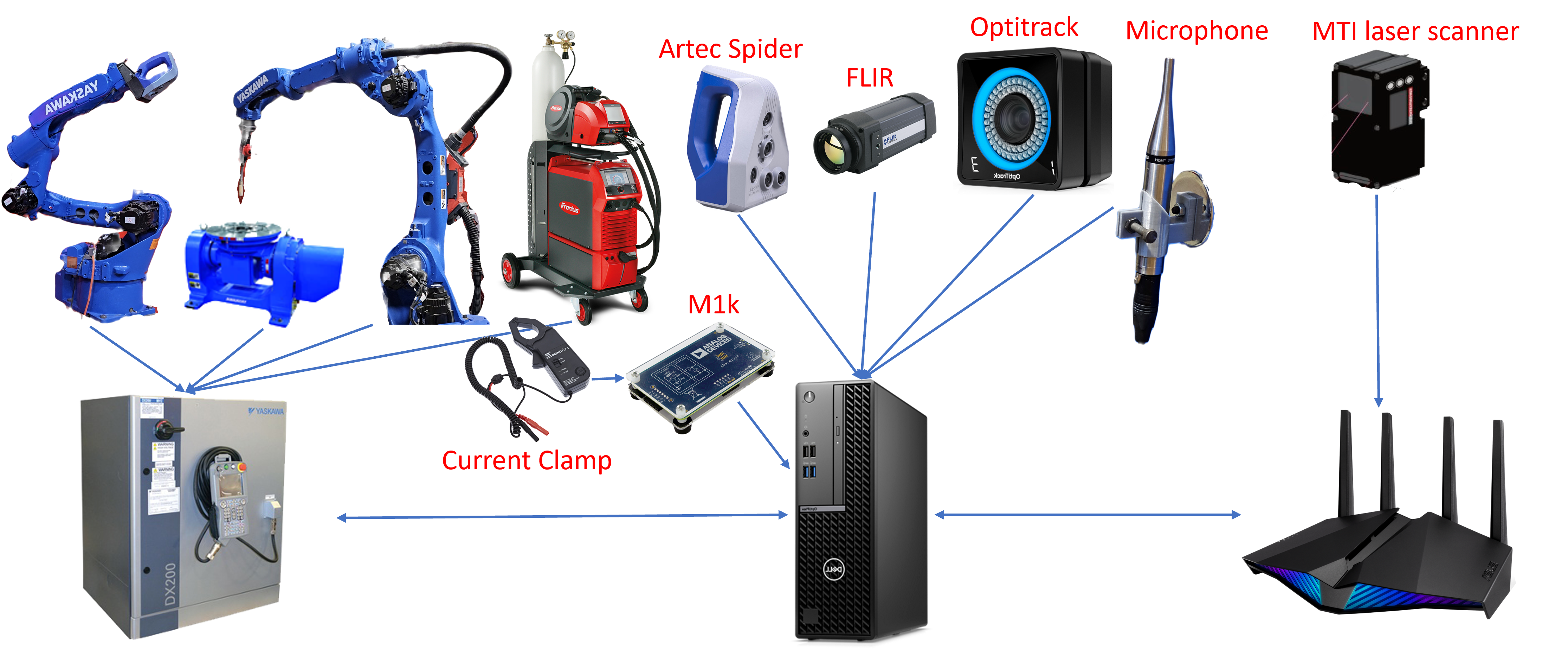}
    \caption{WAAM Cell Hardware Component Architecture.
    }
    \label{fig:architecture}
\end{figure*}

\begin{figure}
    \centering
    \includegraphics[width=0.9\textwidth]{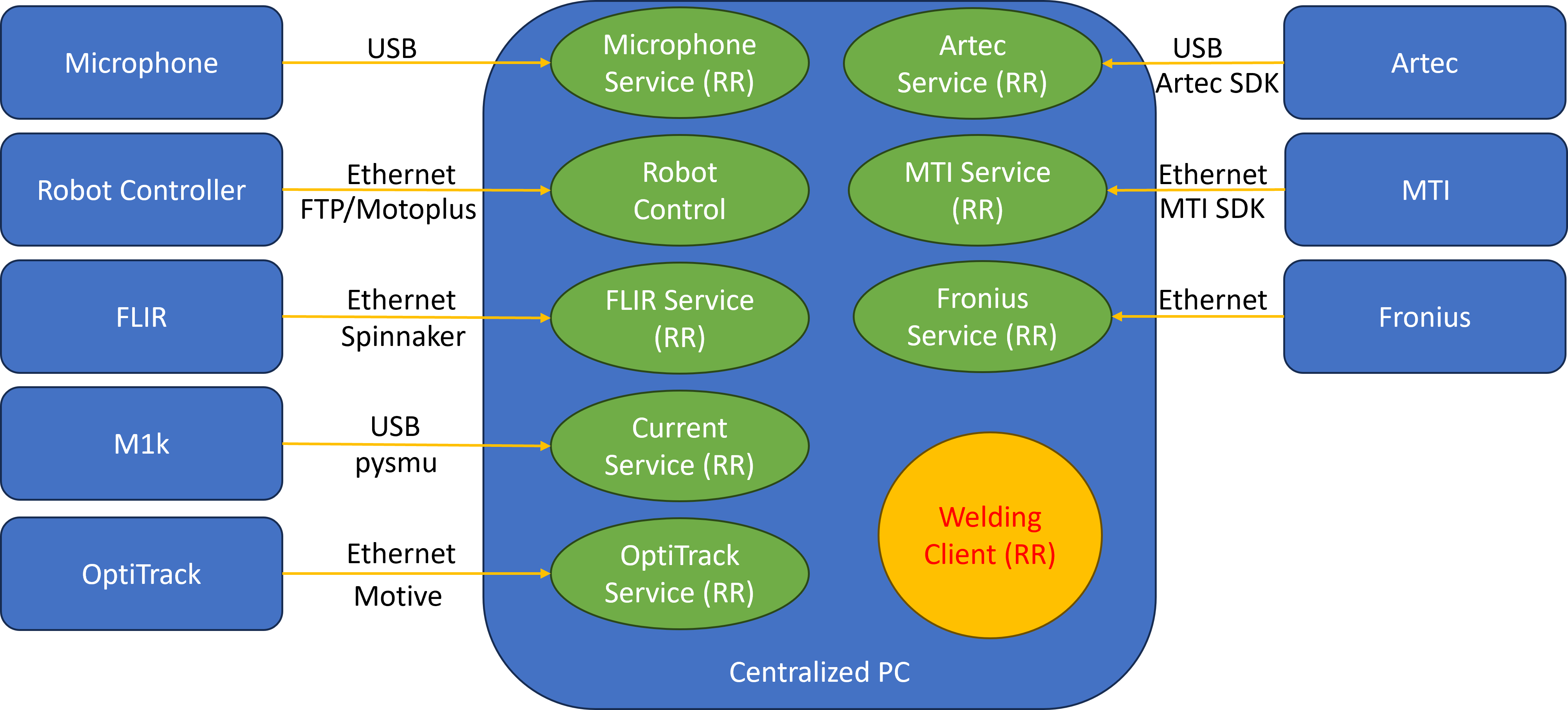}
    \caption{WAAM Cell Software Architecture.
    }
    \label{fig:software_architecture}
\end{figure}

A WAAM cell consists of multiple robots, sensor suite, and a system computer. The robots will include a welding robot, a positioner, and possibly a process monitoring robot.  They typically come in a package including integrated robot and weld controllers.  The in-process sensor suite may include IR camera for temperature monitoring, microphone for acoustic input, weld current sensor, laser line scanner for layer height/width.  There may also be measurement systems for robot calibration, such as a motion capture system, and 3D scanner for post-printing metrology. The system computer is responsible for motion programming, data acquisition, and user interface.  Fig.~\ref{fig:architecture} shows the main hardware components used in our testbed.
Fig.~\ref{fig:software_architecture} shows the software architecture that integrates these components together allowing them to exchange data and commands. 
We use Robot Raconteur (RR) \cite{RR1} as the process middleware due to its versatility (supporting multiple programming languages and operating systems), ease of implementation, and low latency.  
RR wraps each API library into an RR driver which can function as a standalone service node to exchange data with the device and share the data with other RR nodes.  These nodes run concurrently on the process PC.  The user can use a client node (shown as the welding client) to connect to all other RR services to interact with external devices in the system. The component interface drivers in our testbed are listed in Table~\ref{table:software_components}.
\begin{table}[!ht]
\centering
\begin{tabular}{|c|c|c|c|c|} 
 \hline
 Driver & I/F & Protocol & API & Rate \\ 
 \hline\hline
 Microphone \cite{rr_microphone}        &   USB         &   USB    & sounddevice \cite{sounddevice} & 44~KHz     \\
 Robot \cite{dx200_motion_progam_exec}  &   Ethernet    &   UDP    & DX200 API \cite{dx200_ethernet}& 250 Hz   \\   
 FLIR \cite{rr_flir}                    &   Ethernet    &   UDP    & Spinnaker \cite{spinnaker}     & 30 FPS     \\
 M1k \cite{rr_m1k}                      &   USB         &   USB    & pysmu \cite{pysmu}             & 1000 Hz   \\ 
 OptiTrack \cite{rr_optitrack}          &   Ethernet    &   UDP    & Motive \cite{motive}           & 240 Hz   \\
 Artec \cite{rr_artec}                  &   USB         &   USB    & Artec SDK \cite{artec_sdk}     & 7 FPS   \\
 MTI \cite{rr_mti}                      &   Ethernet    &   TCP    & MTI SDK \cite{mti_sdk}         & 140 Hz   \\
 Fronius \cite{rr_fronius}              &   Ethernet    &   Modbus & through DX200                            & 10 Hz   \\
 \hline
\end{tabular}
\caption{Software Component Details}
\label{table:software_components}
\end{table}

\section{Algorithm Design}
\label{sec:algorithm}

\subsection{Uniform Non-Planar 3D Slicing} 

%
%

Printing of any 3D part requires slicing, which converts the 3D model into a series of 2D layers.
A key challenge for complex geometries is overhang structures and unsupported spans.  Using support structures \cite{overhang} is challenging for WAAM due to the liquid molten metal.
%
%
While there are generic support-free slicing algorithms \cite{support_free_slicing} using voxel representation that has been applied to the WAAM process \cite{waam_slicing}, the computation time could be excessive for complex shapes. 
We use a simple and efficient projection algorithm based on the observation that the added degree of freedom of the positioner allows slicing to be along the surface of the desired curve. The algorithm below describes our slicing method for a 3D surface shape (such as a wall, blade, or cup). 
\newtheorem{algorithm}{Algorithm}
\def\calC{{\cal C}}
\begin{algorithm}
Given a desired surface $\calC$ on the base plate in the positioner frame.  Denote the 1D curve from the intersection between $\calC$ and the base plate as layer 0.  Sample layer 0 uniformly to a set of point $\{p_0^j\}$, $j=1,\ldots,N$.  Iterate the layer index $i=0,1,\ldots$ until the termination condition is reached.
    \begin{itemize}
    \item Let $p_i^j$ denote the $j$th point on the 1D curve describing layer $i$. Generate the corresponding point on the next layer using 
    \begin{equation}
        p_{i+1}^j = \mbox{Proj}(p_{i}^j + h \, a_i^j)
        \label{eq:slicing}
    \end{equation}
    where $h$ is a specified height increment, $a_i^j$ is the increment direction, and \mbox{Proj} is the projection onto $\calC$. 
    The increment direction is given by 
    \begin{equation}
     a_i^j=t_i^j \times n_i^j
     \label{eq:increment}
    \end{equation}
    where $t_i^j$ is the path tangent and $n_i^j$ is the outward surface normal. 
    The projection operation returns an empty set if there is the point increment does not intersect with $\calC$. This process is shown in Fig.~\ref{fig:slicing_demo}
    \item The algorithm terminates when the next layer contains no points.    
\end{itemize}
\end{algorithm}
The surface $\calC$ is typically given by a CAD description, consisting of triangular meshes.  The projection operation in \eqref{eq:slicing}, $\mbox{Proj}(p)$ is executed with the following step: 
\begin{itemize}
    \item Find $n$ closest mesh vertices to $p$ ($n$ at this point is manually selected, e.g., for a fine mesh geometry describing the fan blade, we use $n=50$) and fit a plane to these points. 
    \item Apply orthogonal projection of $p$ to the plane.
    \item Check if the projection is within the convex hull of the $n$ vertices \cite{convex_hull}.  If yes, return the projection as  $\mbox{Proj}(p)$, otherwise return an empty set.
\end{itemize}
If the projected point in the middle of the layer returns empty, then the layer is broken into segments. The edge point of each segment is extended along the curve tangent until the edge of $\calC$ is found.  
The same slicing operation is applied to each segment until the termination condition is reached.
Note that $\calC$ is oriented such that the outward normal of the base (layer 0) is on the base plate.  This means $a_0^j$ is pointing opposite of the gravitational acceleration.

\begin{figure}[h]
     \centering
     \begin{subfigure}[b]{0.4\textwidth}
         \centering
         \includegraphics[width=\textwidth]{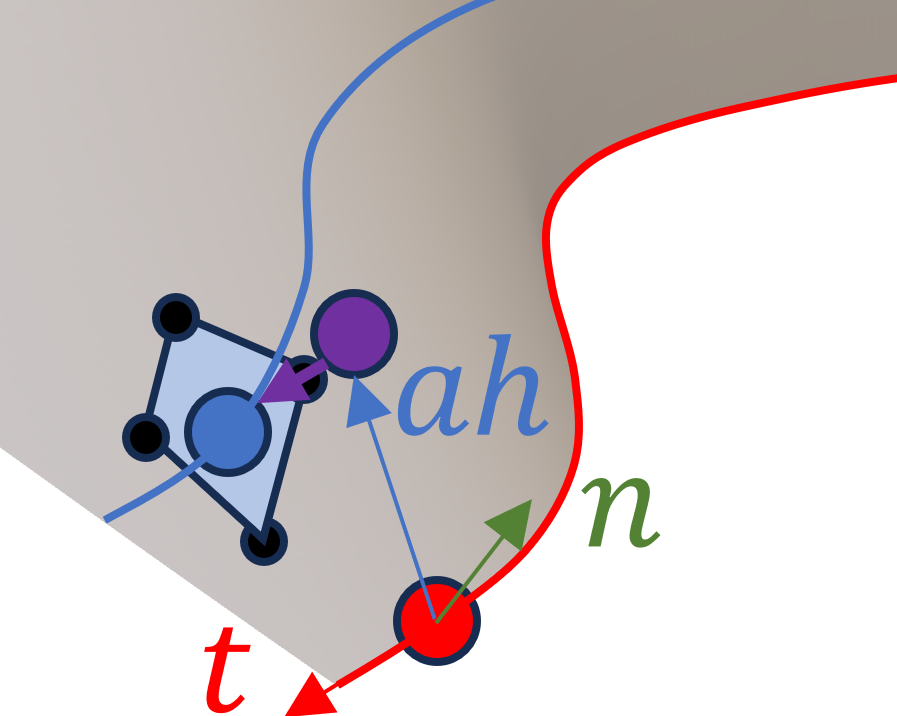}
         \caption{Slicing demonstration for a single point on the curve.}
         \label{fig:slicing_demo}
     \end{subfigure}
     \begin{subfigure}[b]{0.5\textwidth}
         \centering
         \includegraphics[width=\textwidth]{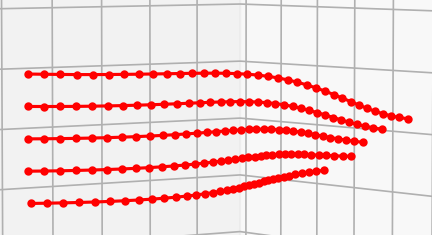}
         \caption{Blade first few slices displayed at $h$ = 10 mm}
     \end{subfigure}
    \caption{Slicing for a generic turbine blade. The purple point is derived from the red point on the previous layer shifting along $t\times n$ by $h$, and then projected onto a fitted plane formed by n black vertices to get the blue point on the next layer.}        
    \label{fig:slicing}
\end{figure}

Fig.~\ref{fig:slicing} shows the slicing result for a generic turbine blade geometry with $h$=1 mm (the figure shows every tenth slice). The point separation distance in the spatial sampling of the first layer is 0.5 mm (the figure shows every tenth point on each slice). 
For the actual WAAM printing of this part, we use a much smaller height increment of $h$=0.1 mm. This allows us to experiment with different deposition rates (and hence different heights) by skipping slices with finer resolution without re-slicing.


\begin{figure}[H]
    \centering
    \begin{subfigure}[b]{0.45\textwidth}
        \centering
        \includegraphics[width=\textwidth]{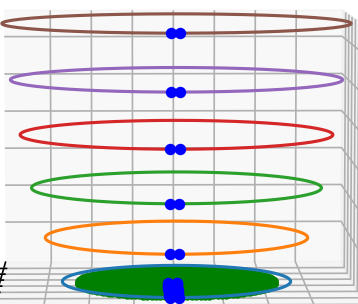}
        \caption{Cup Geometry Slicing}
        \label{fig:cup_slice}
    \end{subfigure}
    \hfill
    \begin{subfigure}[b]{0.45\textwidth}
        \centering
        \includegraphics[width=\textwidth]{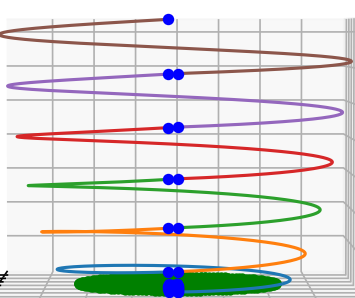}
        \caption{Warped Layers of Slicing}
        \label{fig:cup_spiral}
    \end{subfigure}
    \caption{Layer Warping of Sliced Layers of a Cup Geometry. Connection points are highlighted as blue dots, and all the warped layers essentially form a continuous spiral motion.}
    \label{fig:warping}
\end{figure}

If the part has an analytical description, then we can compute the increment direction in \eqref{eq:increment} directly to obtain the next layer, with layer separation along the increment direction.  This is particularly simple for an axisymmetric part such as a cup, funnel, or bell, where the part consists of concentric circles of varying radii, as shown in Fig.~\ref{fig:warping}(a).


Major defects of the WAAM process occur at the start and end of a bead, namely the arc-ignition and arc-extinguishing point. Studies have shown fewer arc-ignition and arc-extinguishing points could help with stabilizing the rates of deposition to achieve consistent bead width and height \cite{quality_control}. We warp the slice $i$ to slice $i+1$ by applying a scaling factor to each point on the layer:
\begin{equation}
    {p_{i+1}^{j'}} = \alpha_j p_{i+1}^j + (1-\alpha_j) p_i^j 
\end{equation}
where $\alpha_j$ varies linearly from 0 to 1 linearly in the path length: $\alpha = \lambda_j/\lambda_f$, $\lambda$ is the path length of $p_i^j$ on layer $i$ and $\lambda_f$ is the length of layer $i$.
This allows each layer to connect the end of the previous layer and the start of the next layer as shown in Fig.~\ref{fig:warping}(b) such that the robot motion could be continuous without turning arc on and off.

\subsection{Positioner Gravity Alignment}

\begin{figure}[ht]
    \centering
    \includegraphics[height=2.0in]{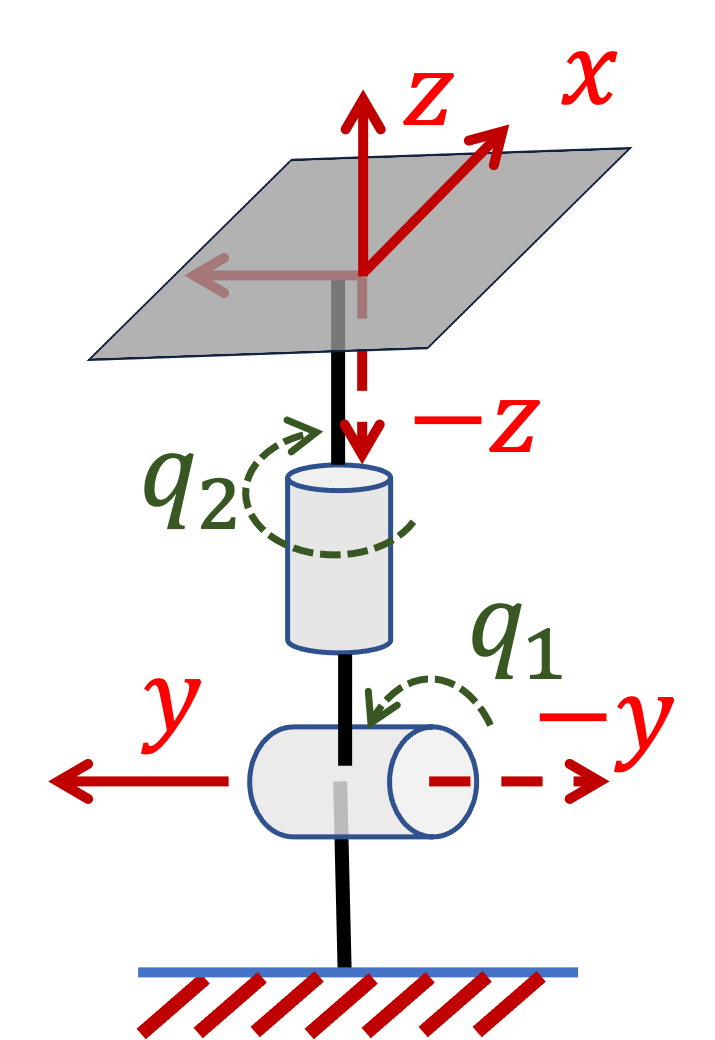}\\
    \includegraphics[height=2.2in]{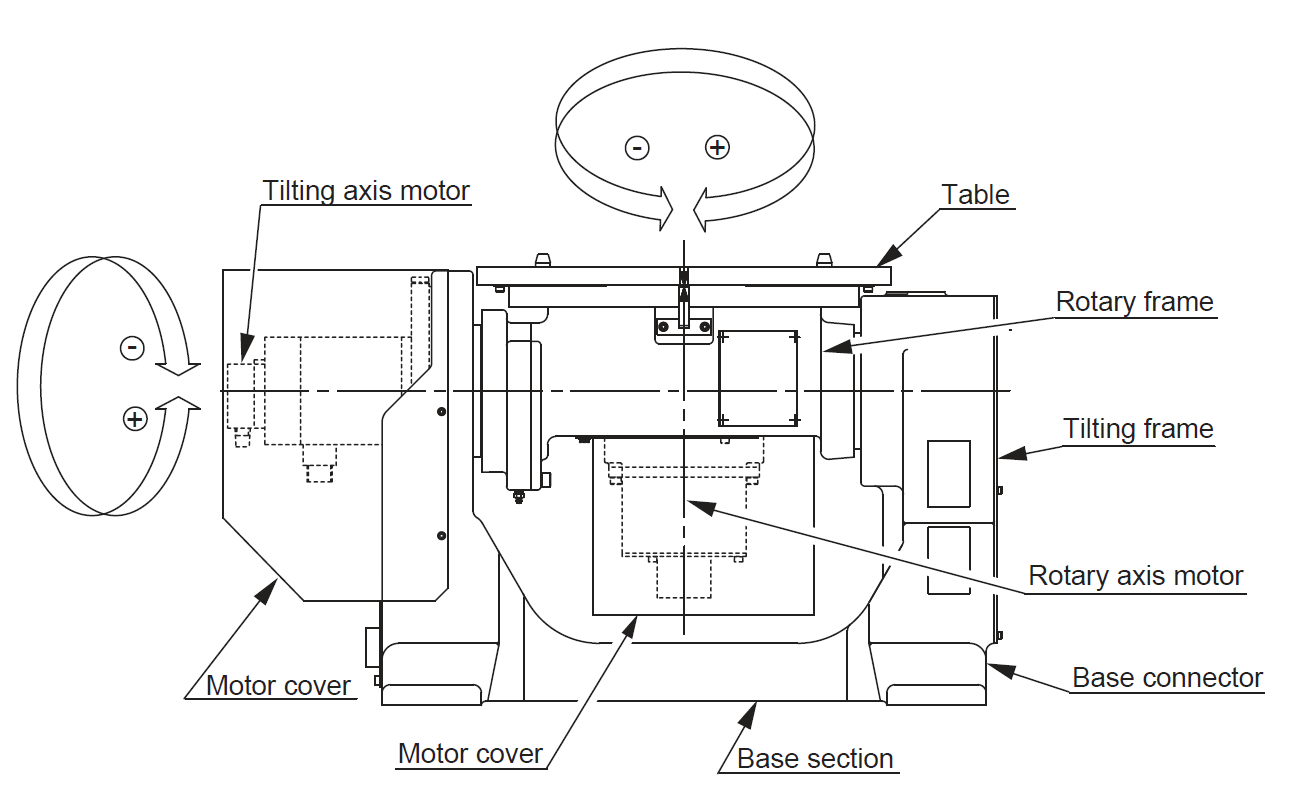}
    \caption{Positioner kinematics in zero configuration \cite{d500b_manual}.}
    \label{fig:basekinematics}
\end{figure}

The molten metal tends to flow along gravity.  To avoid compromising part accuracy, we need to ensure the increment direction $a_i^j$ in \eqref{eq:increment}  is aligned with the gravity. This is achieved by using the positioner. 
The positioner has two rotational joints, where $q_1 \in [-95^{\circ}, 95^{\circ}]$ and $q_2 \in [-72636^{\circ}, 72636^{\circ}]$.  At the zero configuration (when both joint angles are zero), the first axis is in the $-y$ direction and the second axis is in the $z$ direction, as shown in Fig.~\ref{fig:basekinematics}.  The orientation of the base plate in the world frame is given by
\begin{equation}
    R_{0B}= R_{01}R_{12} = \rot(-y,q_1)\rot(-z,q_2)
\end{equation}
where $y=[0,1,0]^T$, $z=[0,0,1]^T$, and $\rot(v,q)= e^{v^\times q}$ denotes the rotation about the unit vector $v$ over an angle $q$ ($v^\times$ is the $3\times 3$ skew-symmetric matrix representing the cross product operation).
Let the increment direction at a given point on a layer be $a=[a_x,a_y,a_z]^T$, as in \eqref{eq:increment}, in the world frame.  We need to choose $(q_1,q_2)$ so that $a=z$ so the weld deposit will build up the layer instead of flowing in other directions.  This means 
\begin{equation}
    z = R_{0B} a \quad \Rightarrow \quad
    \rot(z,q_2)\rot(y,q_1)z=a.
\end{equation}
There are two solutions in general:
\begin{equation}\!\!
\!\!\begin{bmatrix}
    q_1 \\ q_2
\end{bmatrix}
\!\!=\!\! \begin{bmatrix}
    \mbox{atan2}(\sqrt{a_x\sq+a_y\sq},a_z)\\ \mbox{atan2}(a_y,a_x)
\end{bmatrix}\!,\!\begin{bmatrix}
\!    -\mbox{atan2}(\sqrt{a_x\sq+a_y\sq},a_z)\!\\ \mbox{atan2}(a_y,a_x)+\pi
\end{bmatrix}
\end{equation}
Note the two solutions correspond to the opposite tilting directions about the vertical axis. 
We will choose the solution that does not block the view of the monitoring robot (negative $q_1$ in this case).  
When $a$ is close to $z$ (i.e., pointing straight up), $q_1=0$ and there are infinitely many solutions of $q_2$, i.e., the positioner is in a kinematic singularity.  In this case, we simply choose the last value of $q_2$ before $q_1$ falls below a threshold. 
When $a$ is nearly vertical, noise may result in the positioner crossing singularity, causing large joint motion and inaccuracy of the positioner.
To counter this effect, we smooth the positioner trajectory using an averaging filter near the singular configuration.

\subsection{Motion Segment Coordination}

Robotic WAAM system utilizes the multi-robot coordinated motion capability offered by industrial robot vendors, such as Coordinated Control by Yaskawa Motoman \cite{motoman_coordinated}, MultiMove by ABB \cite{multimove}, MultiArm by FANUC \cite{FANUC_coordinated}. In all these systems, motion is specified by waypoints for each robot, and the synchronization is based on the slowest of all robot motion segments so all segments are executed in the same time period. For our WAAM system, we synchronize the motion of the welding robot, positioner, and monitoring robot.
For high-speed applications, blending at waypoints may compromise path tracking accuracy \cite{icra_singlearm}.  For WAAM applications, the traversal speed is much lower, and the blending effect is minimal.  Hence, we choose uniformly spaced Cartesian waypoints (at $d_r$=5 mm) and linear motion segments (moveL) in the relative pose between robot and the positioner.
The path relative velocity of each motion segment is determined by the welding robot velocity command.  Given the desired relative linear velocity $v_r$ and the welding robot segment distance $d_1$, we choose the welding robot path velocity as $v_1=\frac{d_1}{d_r/v_r}$.


\section{Results and Evaluation}
\label{sec:result}


\subsection{WAAM Demonstration Testbed}
\label{sec:testbed_setup}

\begin{figure}[h]
    \centering
    \includegraphics[width=0.8\textwidth]{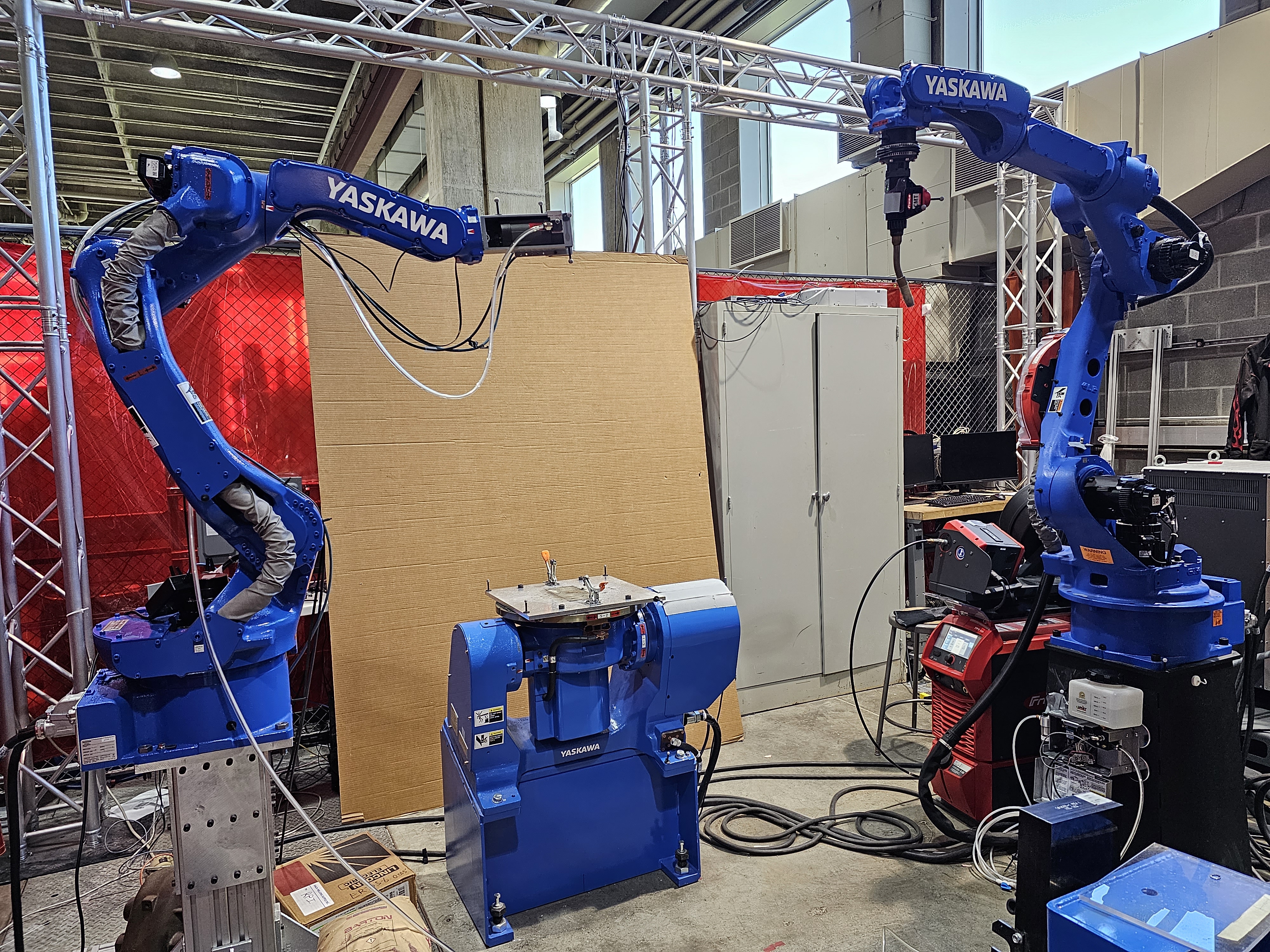}
    \caption{WAAM cell setup, with MA1440-A0 robot on left, D500B in the middle, MA2010-A0 robot on right side and Fronius power source next to MA2010-A0.
    }
    \label{fig:testbed}
\end{figure}

The robotic WAAM architecture described above has been implemented in a demonstration testbed shown in Fig.~\ref{fig:testbed}.  The robotic WAAM cell consists of the following components:
\begin{itemize}
    \item YASKAWA MA2010-A0 6-DoF robot mounted with WF 60i CMT torch
    \item YASKAWA MA1440-A0 6-DoF robot mounted with FLIR A320 camera and MTI laser line scanner
    \item YASKAWA D500B 2-DoF positioner
    \item Fronius TPS500i power source
    \item Earthworks Microphone
    \item Artec Spider 3D Scanner
    \item B\&K Precision CP3 Current Clamp Meter with ADALM1000 (M1k) data acquisition board
    \item OptiTrack Motion Capture System
\end{itemize}
Our welding testbed uses the cold metal transfer (CMT) welding technology based on the Fronius CMT weld controller \cite{fronius}.  CMT is a type of gas metal arc welding that retracts the wire feed when detecting short circuit \cite{CMT}.  This produces low-heat drop-by-drop deposits, resulting in reduced thermal distortions and refined grain structures. Shield gas tank and wire spool are connected to and controlled by the Fronius power source. The positioner holds two clamps to hold aluminum or steel coupons. 

\subsection{Robot Controller Interface}

Yaskawa Motoman offers Coordinated Control to use
two connected DX200 controllers to control the two 6-dof robots with the positioner treated as the external axis.  This setup provides synchronized commands to control all 14 joints using the Motoman robot programming language INFORM \cite{inform}.  To facilitate user motion programming, for Motoman as well as other industrial robots, we developed a universal robot motion command parser.  This parser translates a script of standard motion primitives (moveL, moveC, moveJ) and waypoint specifications to motion commands in specific programming languages, including INFORM for Yaskawa Motoman \cite{inform}, RAPID for ABB \cite{RAPID_manual}, and Karel for FANUC \cite{KAREL_manual}.

The robot program is then sent to the vendor-specific robot controller (or simulator) for execution.  
For our testbed, the INFORM job generated from our parser is transmitted to DX200 using
its Ethernet interface  \cite{dx200_ethernet} to upload and execute the job through the file transfer protocol (FTP). Joint angle feedback is retrieved through a separate MotoPlus program \cite{motoplus} and broadcasted with User Datagram Protocol (UDP) at 250~Hz. 

The relative pose between the robots is calibrated through the OptiTrack motion capture system \cite{optitrack} with reflective markers attached to the robots. This calibration is performed once and saved as local calibration files.

\subsection{Motion Visualization}

\begin{figure}[h]
    \centering
    \includegraphics[width=0.8\textwidth]{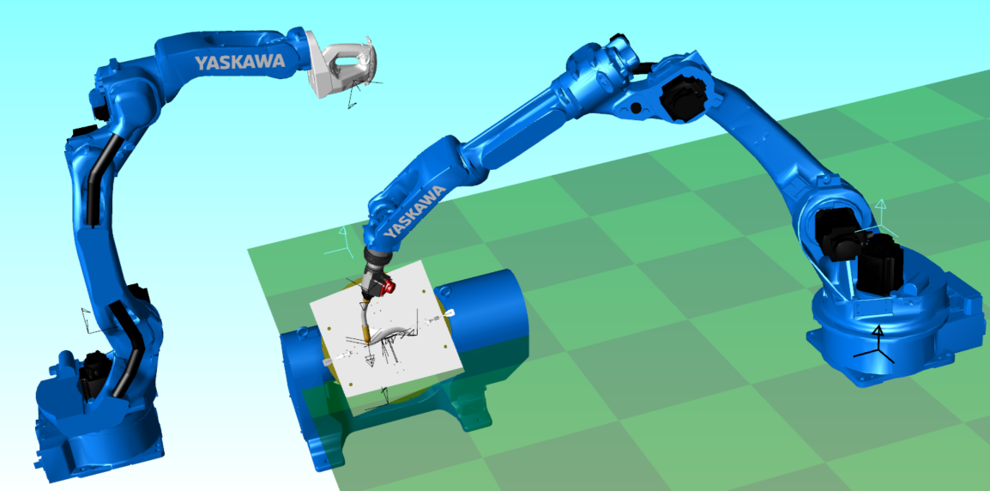}
    \caption{Motosim motion visualization from parser output as an INFORM JOB.}
    \label{fig:motosim}
\end{figure}

Before the actual robot motion execution on the actual robots, it is always prudent to visualize motion in a simulator.  Industrial robot vendors have their own specific simulators, compatible with their specific programming languages and interface standards.  Examples include MotoSim \cite{motosim} for Motoman, RobotStudio \cite{robotstudio} for ABB, and RobotGuide \cite{roboguide} for FANUC. 
Motosim virtual environment is generated from DX200 controller's backup file with identical settings. The output of the parser described above will run in the simulator first (as shown in Fig.~\ref{fig:motosim}) to ensure there is no glaring error in commanded motion.

\subsection{Welding Parameters}
The weld controller, Fronius TPS500i, allows users to specify multiple wire feed rates and save them as separate jobs. Each feed rate has its pre-determined voltage and current combination based on the CMT requirement.  We set up multiple jobs with feed rates in increments of 10 inch-per-minute (ipm).


We experiment with multiple materials to evaluate the range of system performance.  The material type, wire diameter, and the corresponding shield gas, are described below:
\begin{itemize}
    \item Aluminum (ER4043, $\phi$=1.2~mm) with 100\% Argon
    \item Steel Alloy (ER70S-6, $\phi$=0.9~mm) with 75\% Argon, 25\% CO$_{{2}}$
    \item Stainless Steel (ER316L, $\phi$=0.9~mm) with 75\% Argon and 25\% CO$_{{2}}$
\end{itemize}

The two key parameters in any robotic welding applications, including WAAM, are the wire feed rate, $f$, and torch speed $v$ (determined by the robot end effector speed).  The feed rate and wire size determine the rate of deposition in terms of material volume.  The torch speed determines the deposit area and the rate of heat input.  

 The ratio $\alpha=\frac{f}{v}$ is proportionally related to the layer geometry (height and width).  If $\alpha$ is too small, not enough material is deposited and the layer will consists of discrete droplets instead of a smooth surface. If $\alpha$ is too high, there will be uncontrolled buildups.  Slower $v$ means more heat input (proportional to $\frac{VI}{v}$, where $V$ and $I$ are the weld controller voltage and current) and more deposited materials.  Excessive heat will melt the lower level and excessive material will cause uncontrolled buildup.  



\begin{figure}[h]
     \centering
     \begin{subfigure}[b]{0.33\textwidth}
         \centering
         \includegraphics[width=\textwidth]{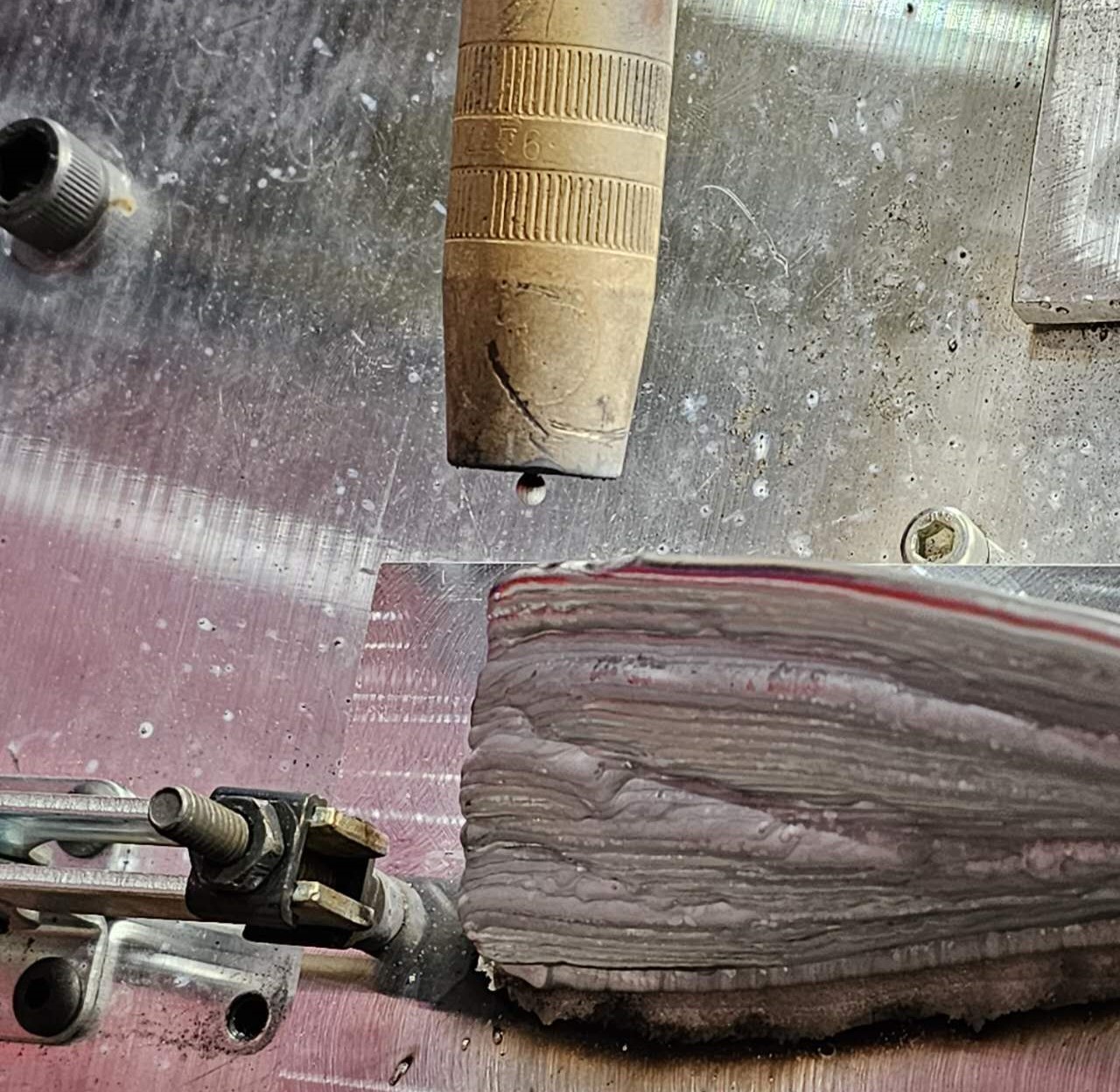}
         \caption{Short Wire}
         \label{fig:short_wire}
     \end{subfigure}
     \begin{subfigure}[b]{0.3\textwidth}
         \centering
         \includegraphics[width=\textwidth]{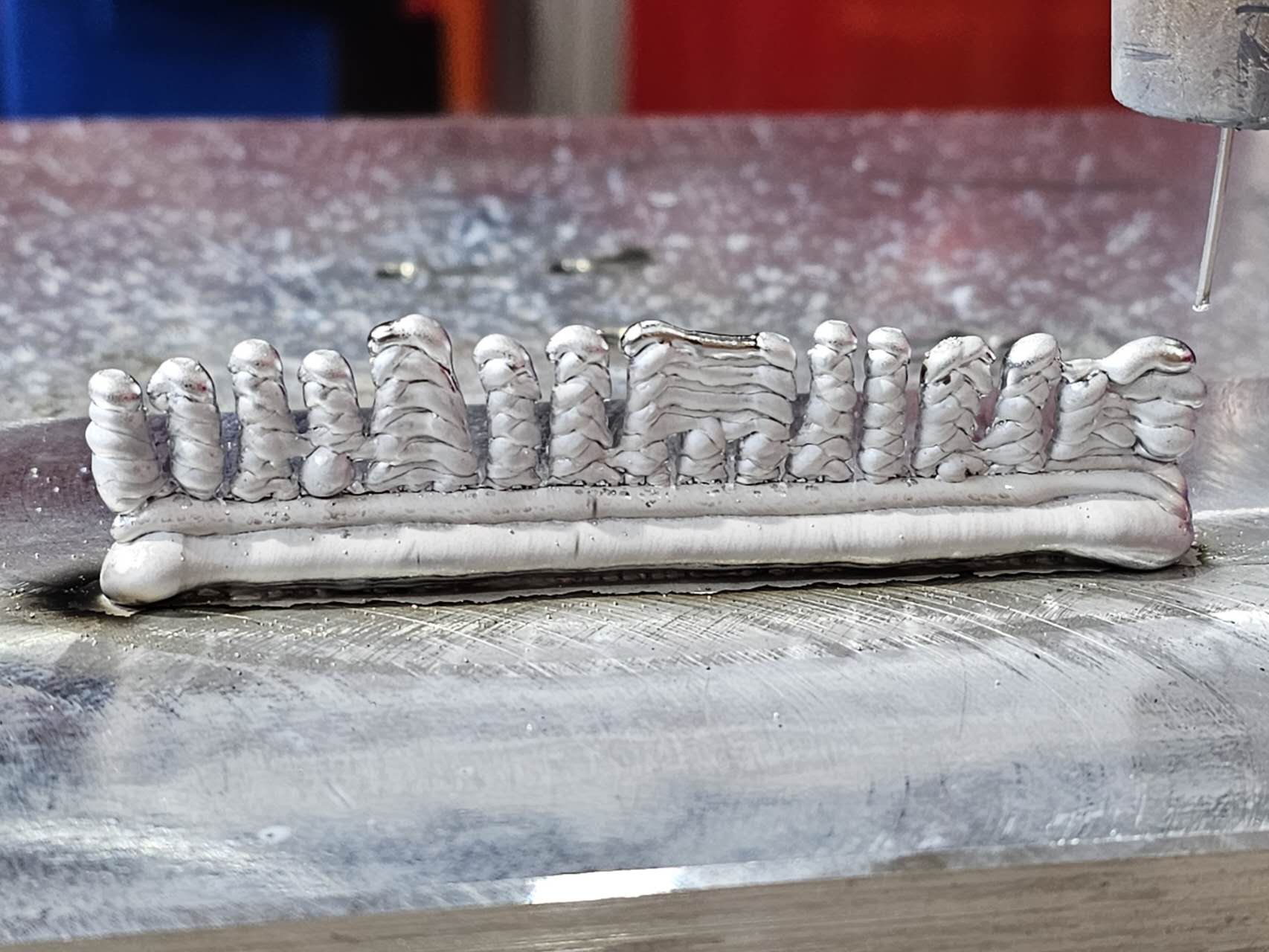}
         \caption{Bubbling}
         \label{fig:bubbling}
     \end{subfigure}
     \begin{subfigure}[b]{0.33\textwidth}
         \centering
         \includegraphics[width=\textwidth]{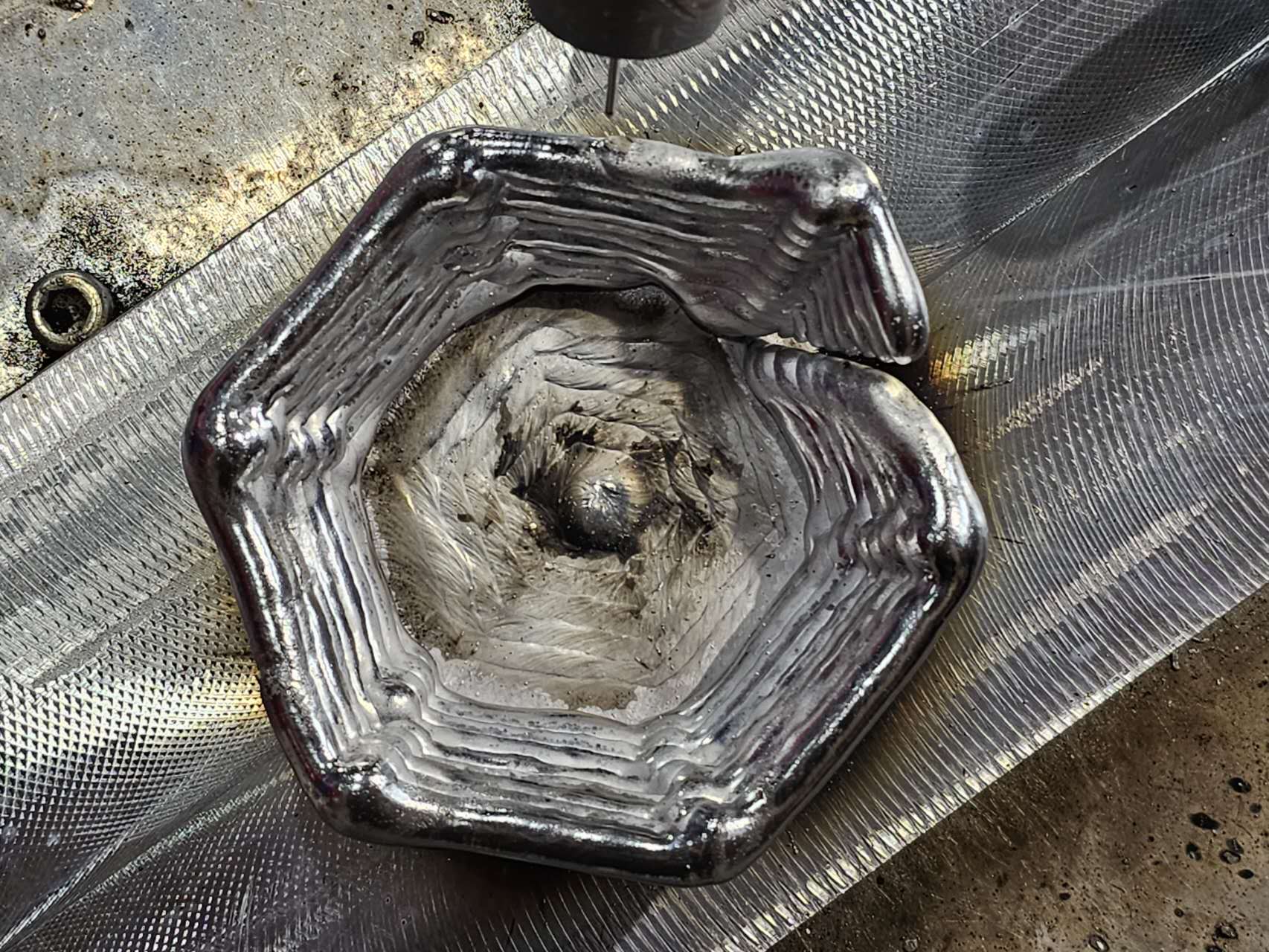}
         \caption{Gap Propagation}
         \label{fig:gap_prop}
     \end{subfigure}
     \begin{subfigure}[b]{0.3\textwidth}
         \centering
         \includegraphics[width=\textwidth]{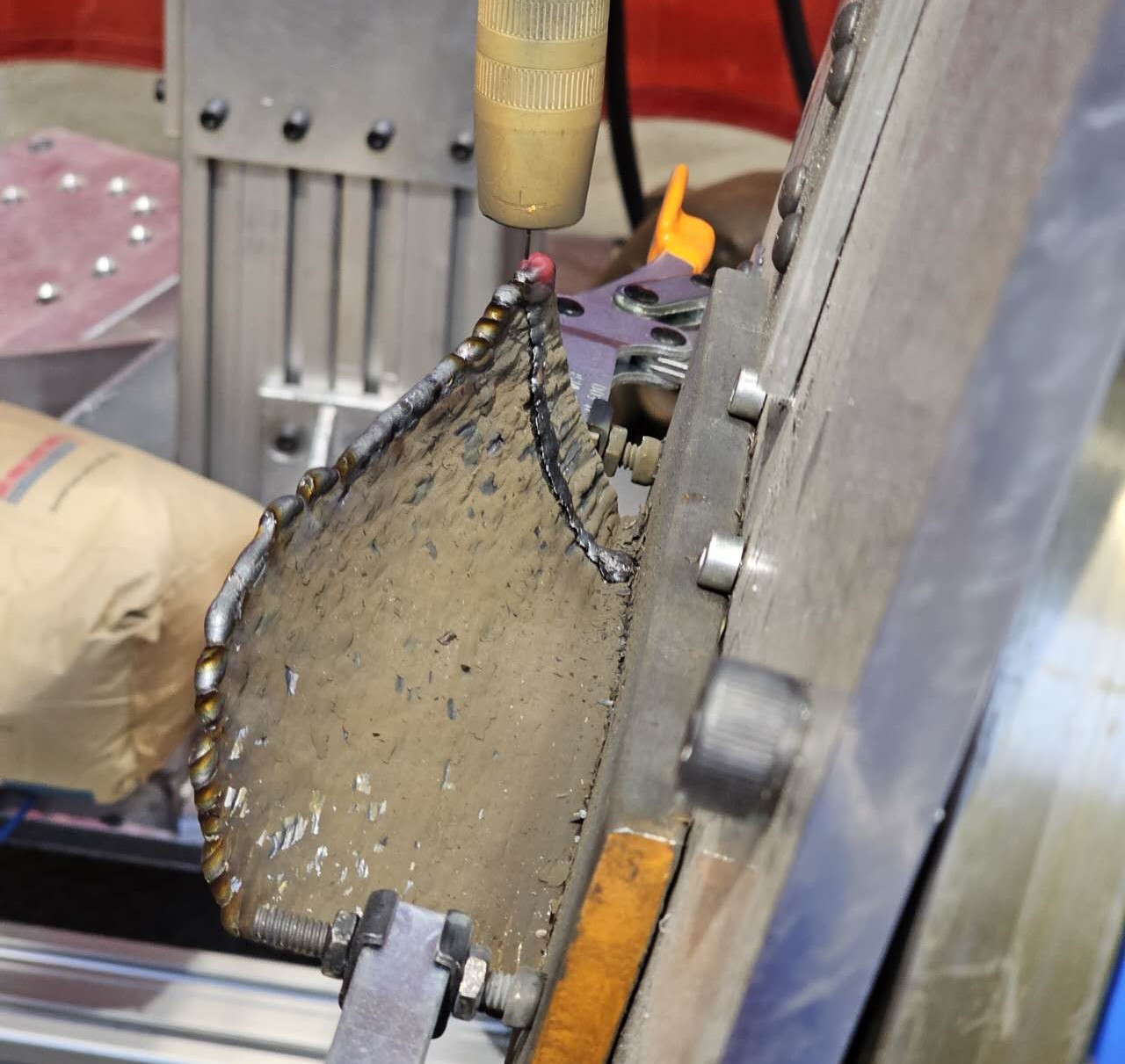}
         \caption{Melting}
         \label{fig:melting}
     \end{subfigure}
     \begin{subfigure}[b]{0.35\textwidth}
         \centering
         \includegraphics[width=\textwidth]{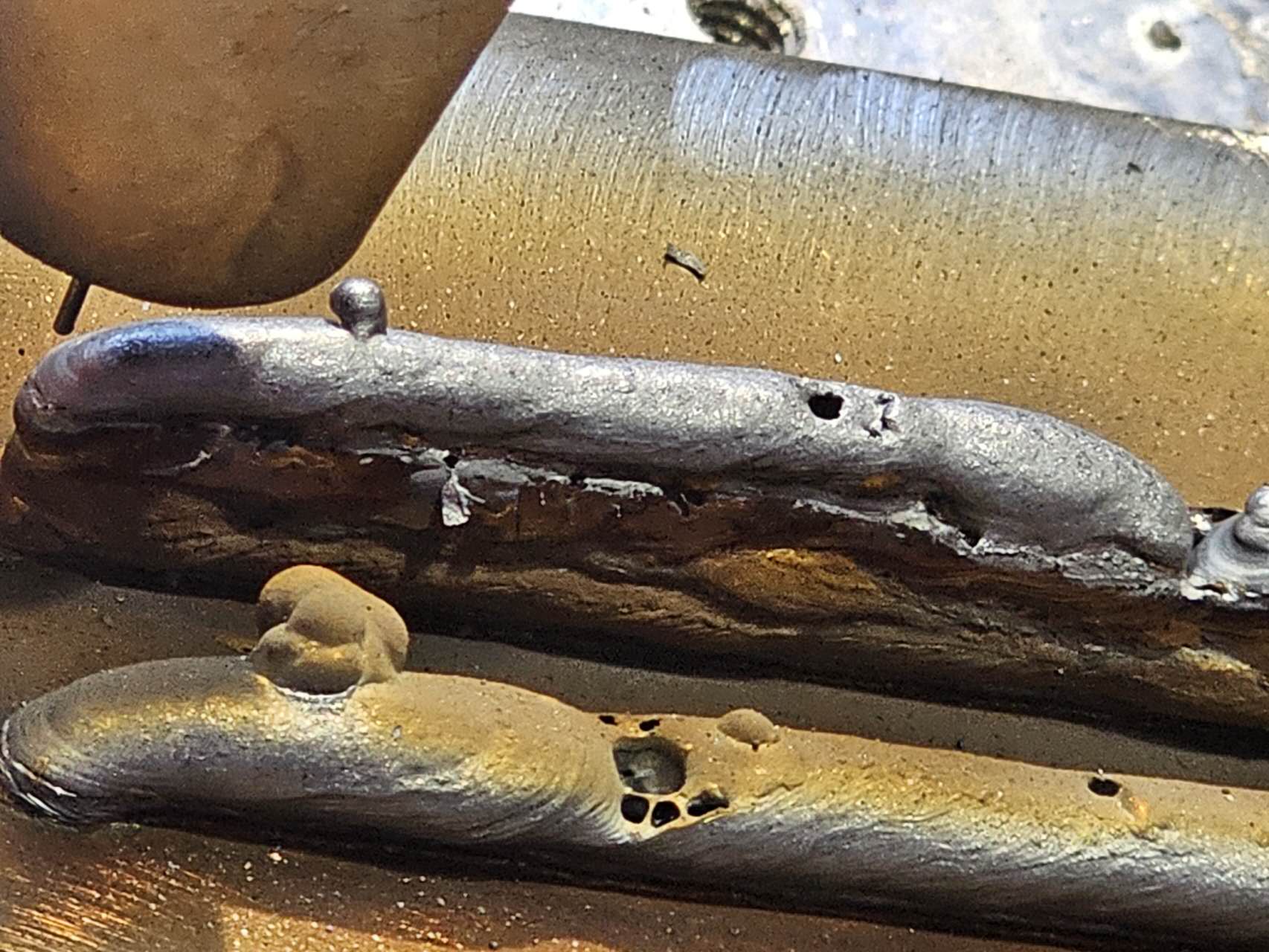}
         \caption{Defect from Shield Gas}
         \label{fig:shield_gas}
     \end{subfigure}
    \caption{Failure Cases for the WAAM Process.}
        
    \label{fig:failure}
\end{figure}

We conducted multiple trial runs to gain insight of the system behavior. 
Fig.~\ref{fig:failure} shows multiple failure cases, such as short metal wire, defect shield gas, undesired melting caused by excessive heat, bubbling and gap propagation due to insufficient materials.
By adjusting $v$ and $f$ using the observed trend, we determine the most suitable combinations:
\begin{itemize}
    \item Aluminum (ER4043, $\phi$=1.2~mm): 9~mm/s, 110~ipm (resulting 1.05~mm deposition height on average)
    \item Steel Alloy (ER70S-6, $\phi$=0.9~mm), 8~mm/s, 100~ipm (resulting in 0.85~mm deposition height on average)
    \item Stainless Steel (ER316L, $\phi$=0.9~mm), 8~mm/s, 130~ipm (resulting in 1.2~mm deposition height on average)
\end{itemize}
Note that the deposition height is slightly different for each material, this is because different materials have different properties like heat transfer rate, melting point, viscosity, etc.  We aim to achieve the best overall formation of the final product to avoid critical failures shown in Fig.~\ref{fig:failure} and the best final product geometric accuracy without human intervention.


\subsection{Demonstration 3D Geometry}
\begin{figure}[h!]
     \centering
     \begin{subfigure}[b]{0.3\textwidth}
         \centering
         \includegraphics[width=\textwidth]{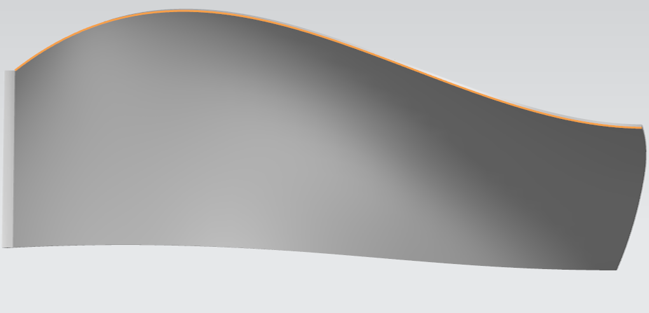}
         \caption{Blade}
         \label{fig:blade}
     \end{subfigure}
     \begin{subfigure}[b]{0.3\textwidth}
         \centering
         \includegraphics[width=\textwidth]{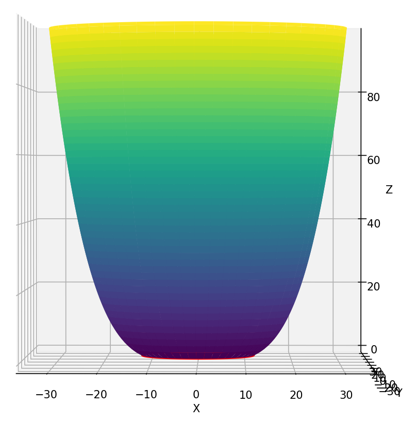}
         \caption{Cup}
         \label{fig:cup}
     \end{subfigure}
     \begin{subfigure}[b]{0.3\textwidth}
         \centering
         \includegraphics[width=\textwidth]{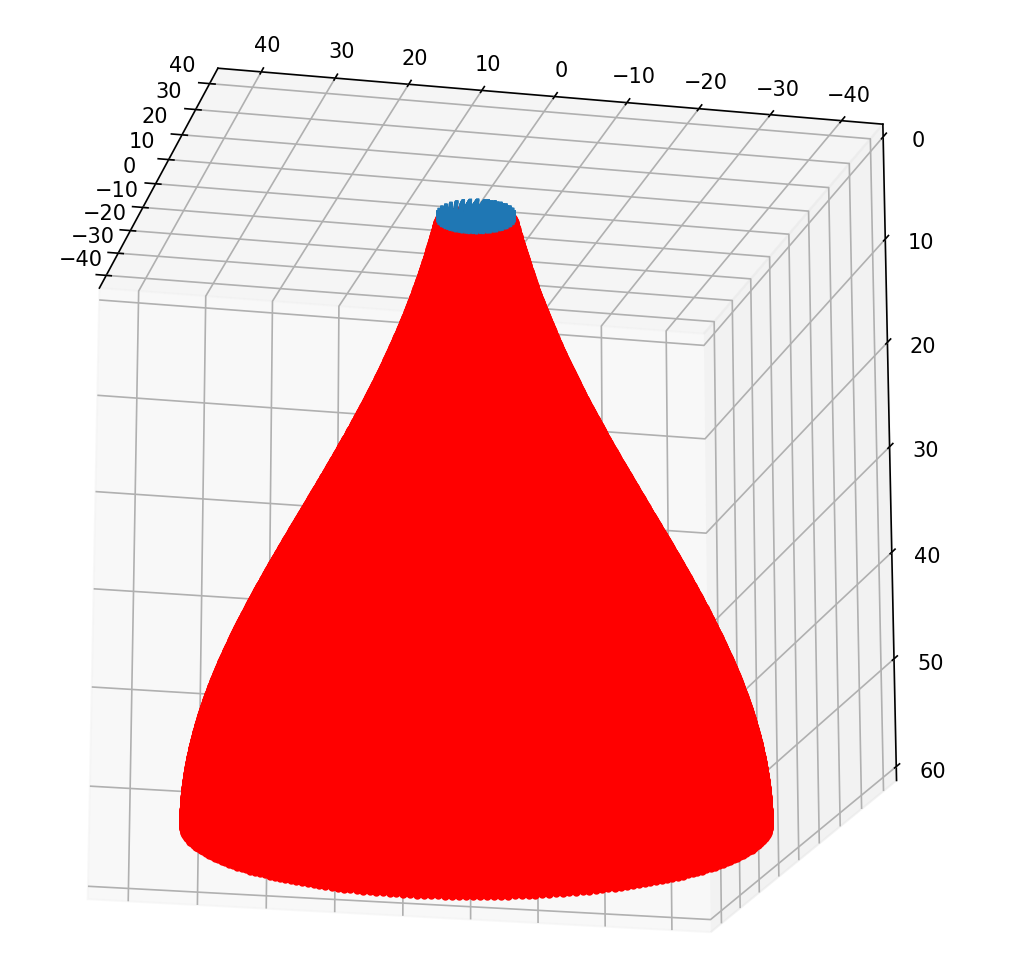}
         \caption{Bell}
         \label{fig:bell}
     \end{subfigure}
     \begin{subfigure}[b]{0.3\textwidth}
         \centering
         \includegraphics[width=\textwidth]{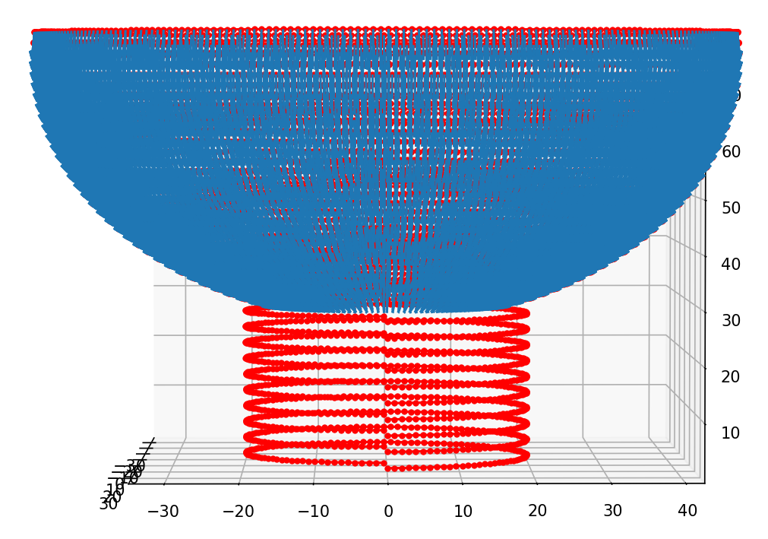}
         \caption{Funnel}
         \label{fig:funnel}
     \end{subfigure}
     \begin{subfigure}[b]{0.3\textwidth}
         \centering
         \includegraphics[width=\textwidth]{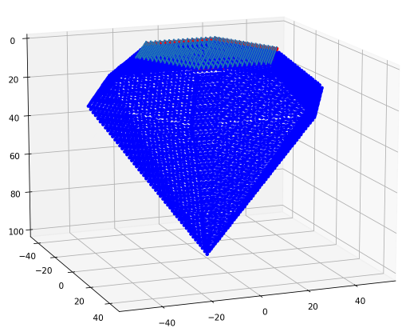}
         \caption{Diamond}
         \label{fig:diamond}
     \end{subfigure}
    \caption{Test Geometries for the WAAM Process.}    
    \label{fig:geometries}
\end{figure}

\begin{figure}[h]
     \centering
     \begin{subfigure}[b]{0.45\textwidth}
         \centering
         \includegraphics[height=1.3in]{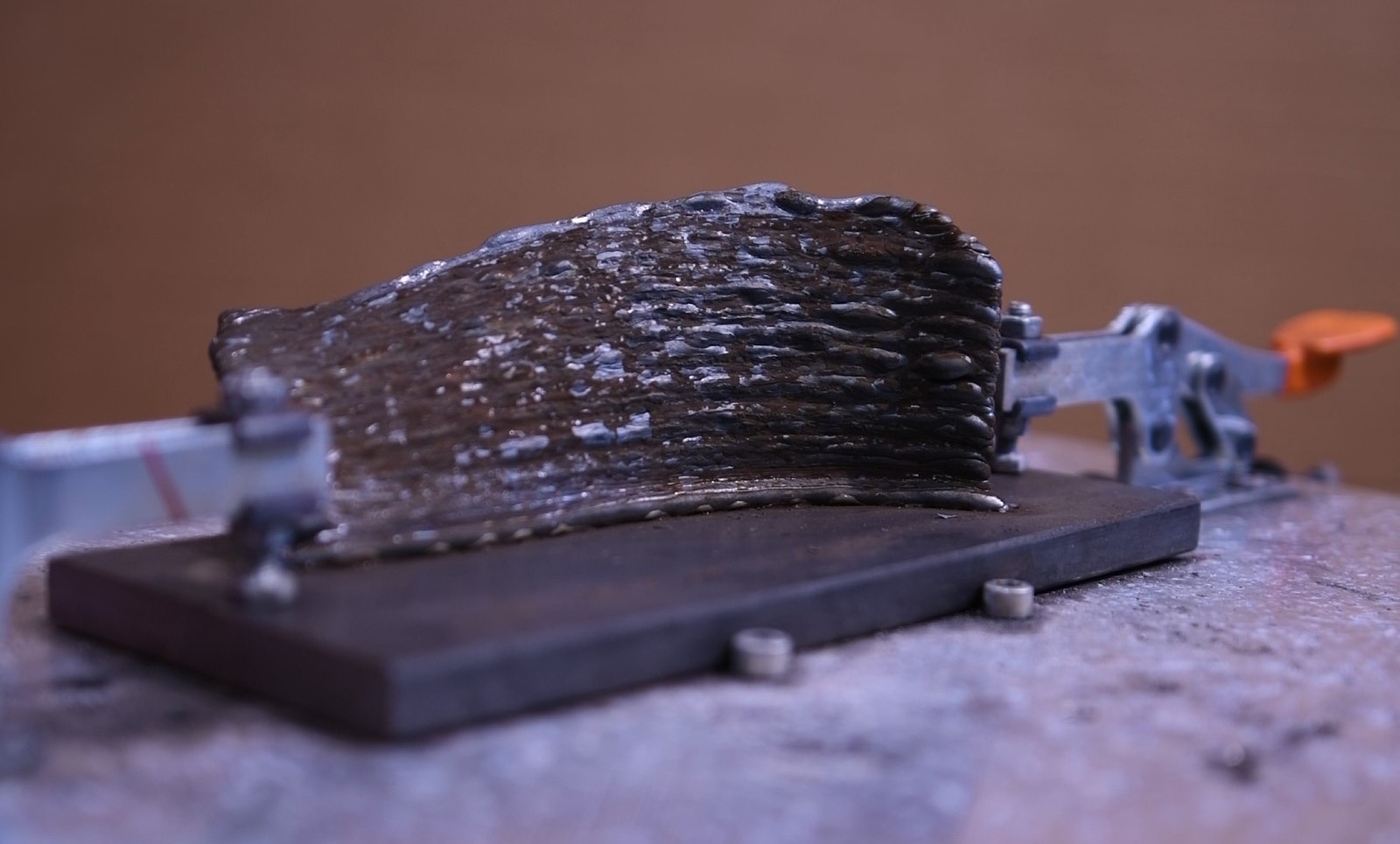}
         \caption{Steel Alloy Blade}
     \end{subfigure}
     \begin{subfigure}[b]{0.45\textwidth}
         \centering
         \includegraphics[height=1.3in]{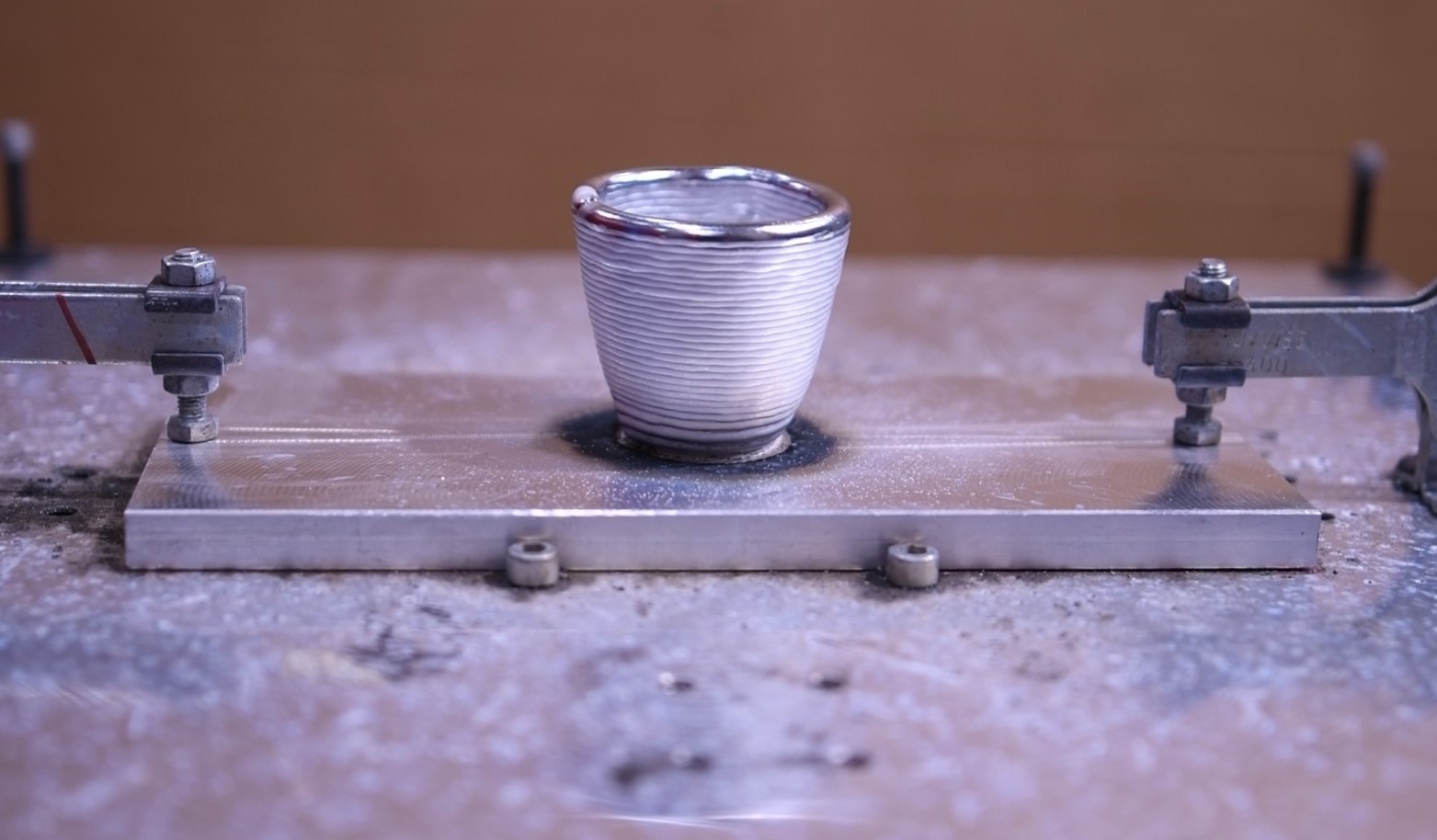}
         \caption{Aluminum Cup}
     \end{subfigure}
     \begin{subfigure}[b]{0.45\textwidth}
         \centering
         \includegraphics[height=1.3in]{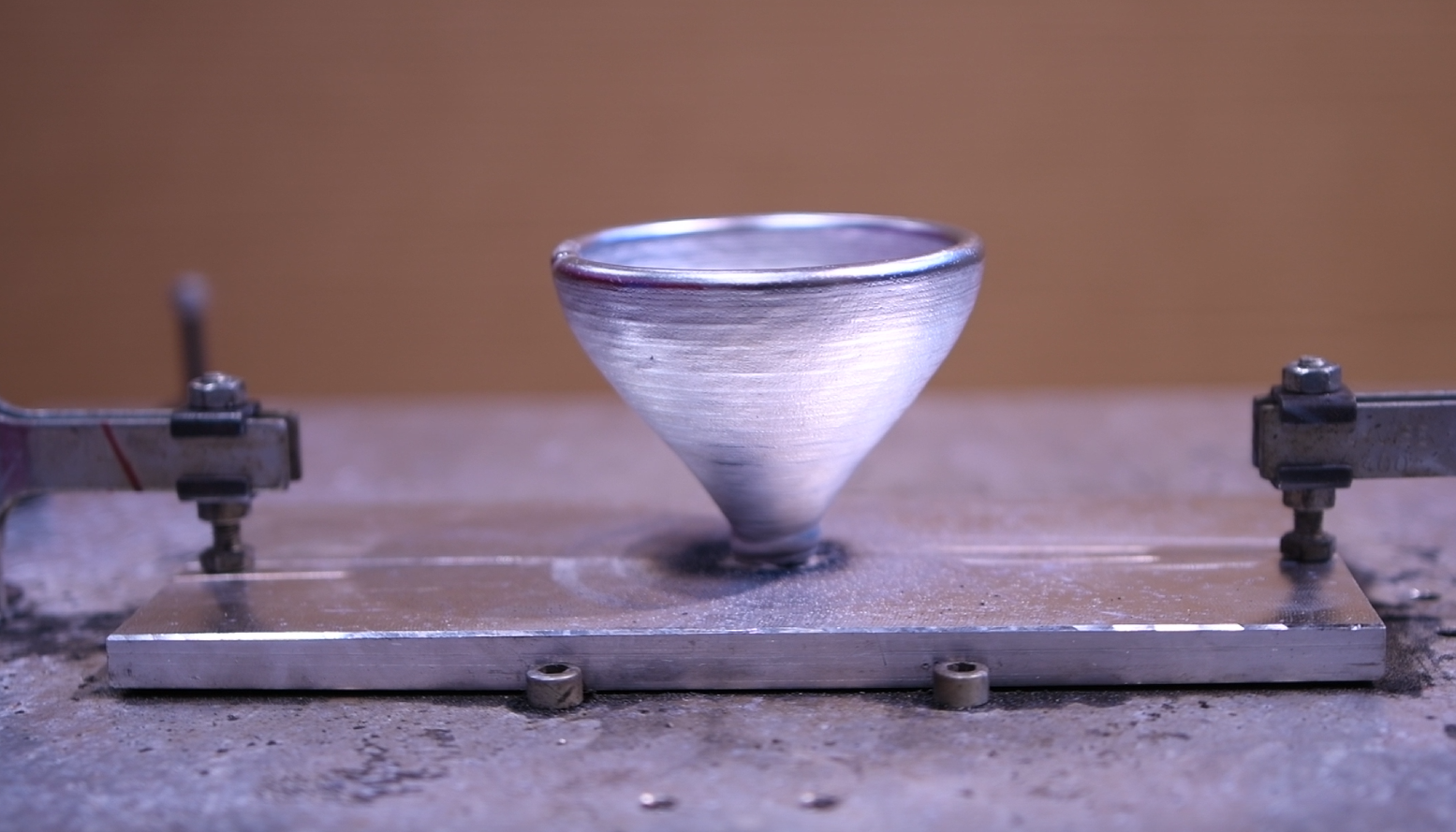}
         \caption{Aluminum Bell}
     \end{subfigure}
     \begin{subfigure}[b]{0.45\textwidth}
         \centering
         \includegraphics[height=1.3in]{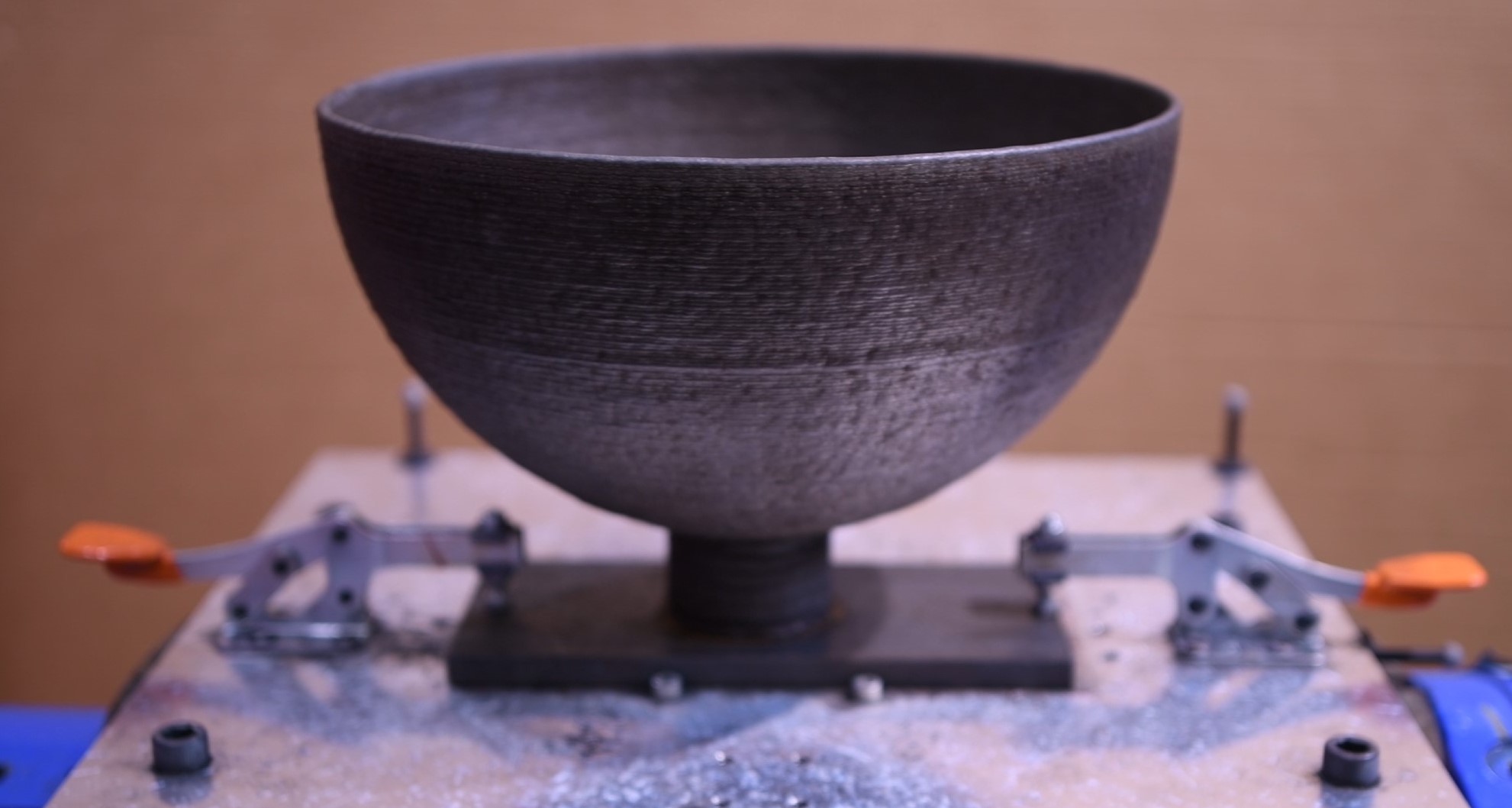}
         \caption{Stainless Steel Funnel}
     \end{subfigure}
     \begin{subfigure}[b]{0.45\textwidth}
         \centering
         \includegraphics[height=1.3in]{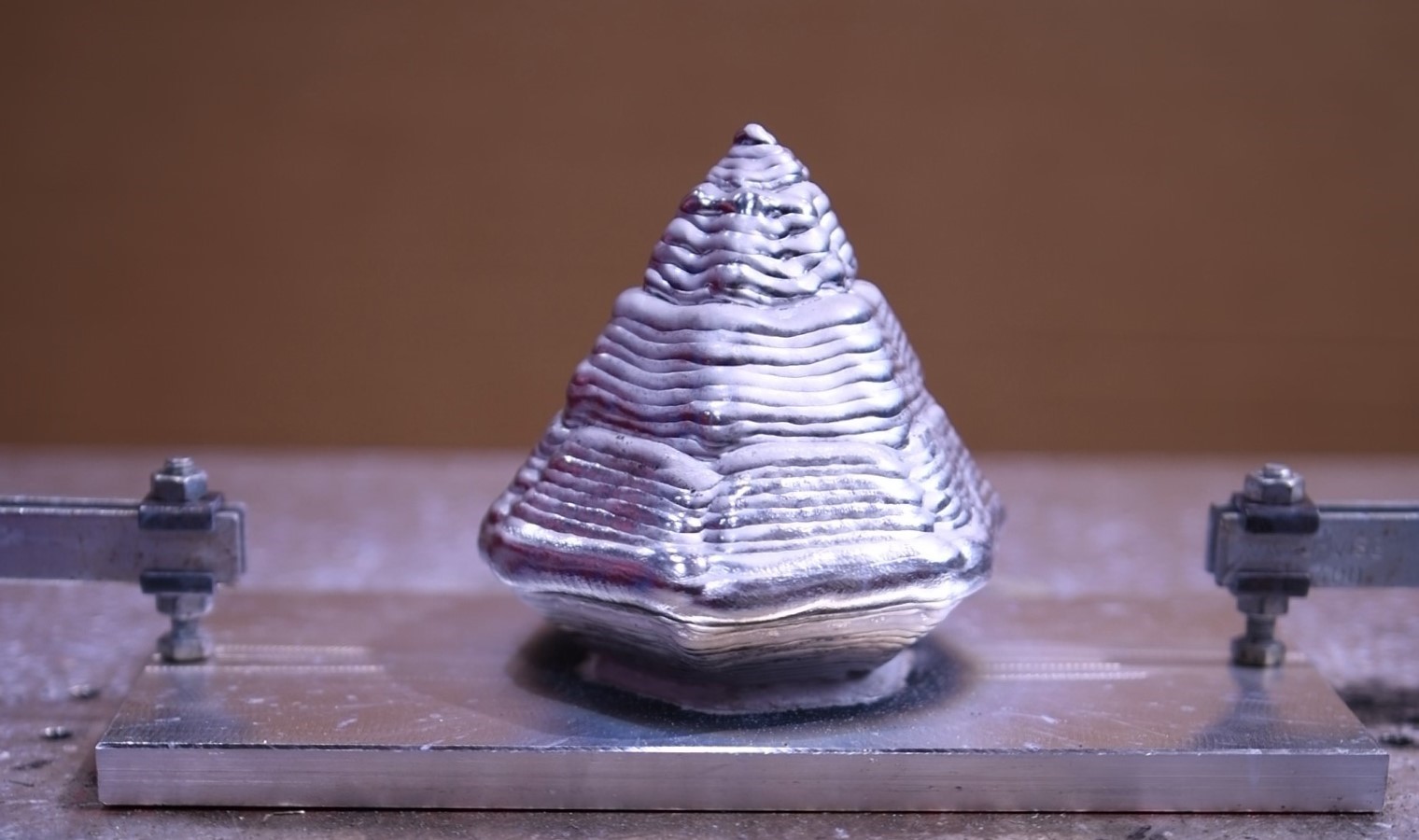}
         \caption{Aluminum Diamond}
     \end{subfigure}
    \caption{WAAM Result Geometries.}
        
    \label{fig:welded_geometries}
\end{figure}

We choose several test geometries for the three types of materials (alumnium, steel alloy, stainless steel) to explore the performance of the robotic WAAM system:
\begin{itemize}
    \item Blade: scaled version of generic turbine blade,  representative of complex 3D geometry
    \item Cup and Bell: spiral closed cycle layers (instead of back and forth)
    \item Funnel: large volume and spiral closed cycle layers
    \item Diamond: Enclosed geometry and sharp edges
\end{itemize}
The CAD models are shown in Fig.~\ref{fig:geometries}, and the WAAM results are shown in Fig.~\ref{fig:welded_geometries}. 

\subsection{Process Monitoring} 

In polymer 3D printing, the layer deposition height is consistent and repeatable for a constant feed rate due to the rapid curing and solidification.  In contrast, the deposition height for each layer in WAAM could vary due to the changing heat transfer conditions to the ambient and to the lower layers. 
An open-loop control with the assumption of a consistent deposition height may result in a short wire and possible collision between the torch and WAAM geometry as shown in Fig.~\ref{fig:short_wire} or long wire and outside of shield gas region.  For complex geometries, the error could also accumulate over the layers.  There has been efforts to control the layer height during printing by using arc current and in-process re-slicing \cite{height_control} but the height measurement is indirect and may be inexact.

\begin{figure}[ht]
 \centering
 \begin{subfigure}[b]{0.2\textwidth}
     \centering
     \includegraphics[height=1.8in]{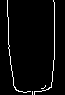}\vspace{.55in}
     \caption{Torch Template}
     \label{fig:torch_template}
 \end{subfigure}
 \qquad
 \begin{subfigure}[b]{0.3\textwidth}
     \centering
     \includegraphics[height=2.5in]{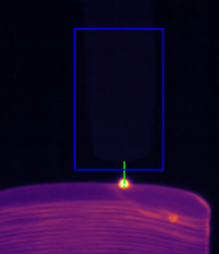}
     \caption{WAAM Wire Tracking}
     \label{fig:ir_wire_track}
 \end{subfigure}
 \caption{Real-time WAAM Wire Length Tracking with 16-bit IR Image.}
 \label{fig:wire_tracking}
\end{figure}

We have a FLIR camera mounted on the separate 6-dof robot to monitor the process.  The robot end-effector motion is commanded to aim the camera at the welding torch.  We use the RR service based on the manufacture's interface, Spinnaker SDK \cite{spinnaker}, to capture the IR image frames at 30~Hz.  These images may be converted to temperature readings using calibrated emissivity and conversion parameters. Fig.~\ref{fig:ir_wire_track} shows an IR image of the aluminum blade WAAM process. 
There are numerous possible uses of the IR video streams during welding, such as to adjust motion to achieve uniform temperature for consistent geometry and microstructures.  In this paper, we will just present one novel use of the IR images -- using the image to track the standoff wire length and adaptively choose the best-sliced layer from our densely sliced (0.1~mm increment) layers from our slicer. 
We first generate an edge template for the torch, and then use template matching during run-time to identify the torch in the IR image frame. The flame is identified through thresholded connected component labeling, allowing us to determine the layer height and adjust the slice for the next layer. This process is shown in Fig.~\ref{fig:wire_tracking}.

\begin{figure}[h]
    \centering
    \includegraphics[width=0.88\textwidth]{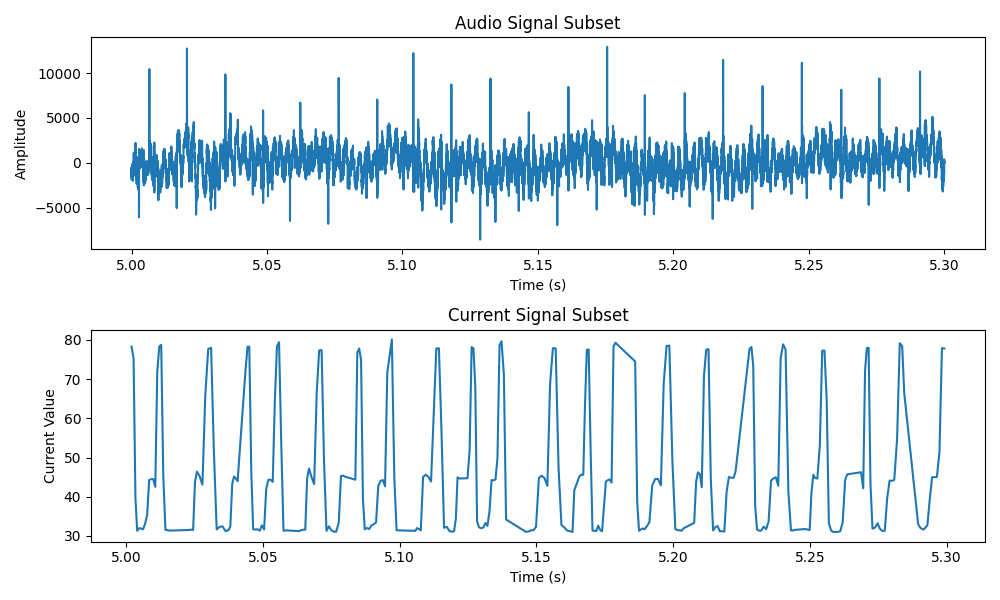}
    \caption{Audio and Current Recording of CMT Welding Process.}
    \label{fig:audio_current}
\end{figure}

Our WAAM architecture also provides RR services to interface to a microphone and a current meter to monitor potential faults in the WAAM process.   The microphone records acoustic signature during the WAAM process as shown in Fig.~\ref{fig:audio_current}. The Fronius interface provides the current reading at 10~Hz, but it is too slow for diagnostics use.  We use a Hall Effect current clamp meter \cite{bk} to measure across the ground cable to provide weld current measurements at 1~KHz as shown in Fig.~\ref{fig:audio_current}.

\subsection{Evaluation}

\begin{figure}[h]
    \begin{subfigure}[b]{0.22\textwidth}
     \centering
     \includegraphics[width=\textwidth]{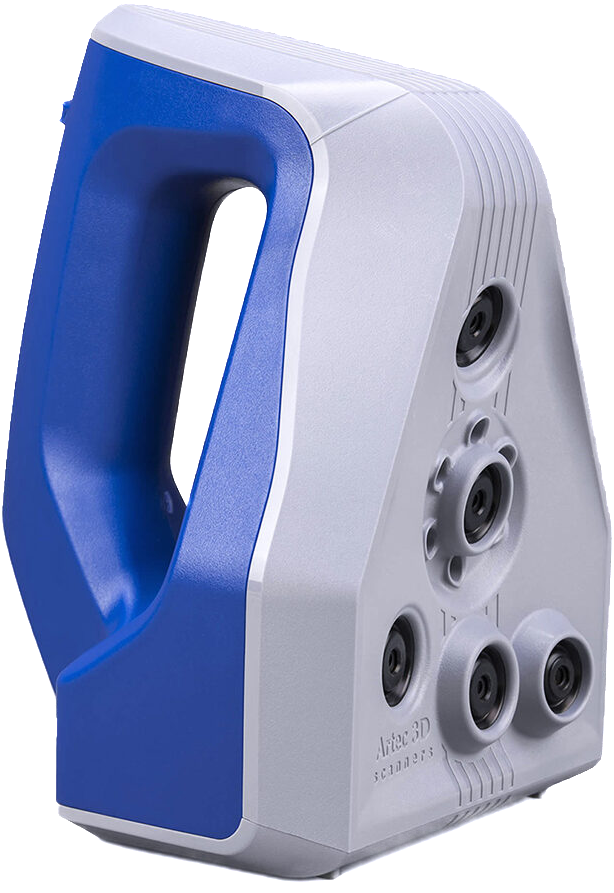}
     \caption{Artec Spider 3D Scanner}
    \end{subfigure}
    \begin{subfigure}[b]{0.55\textwidth}
     \centering
     \includegraphics[width=\textwidth]{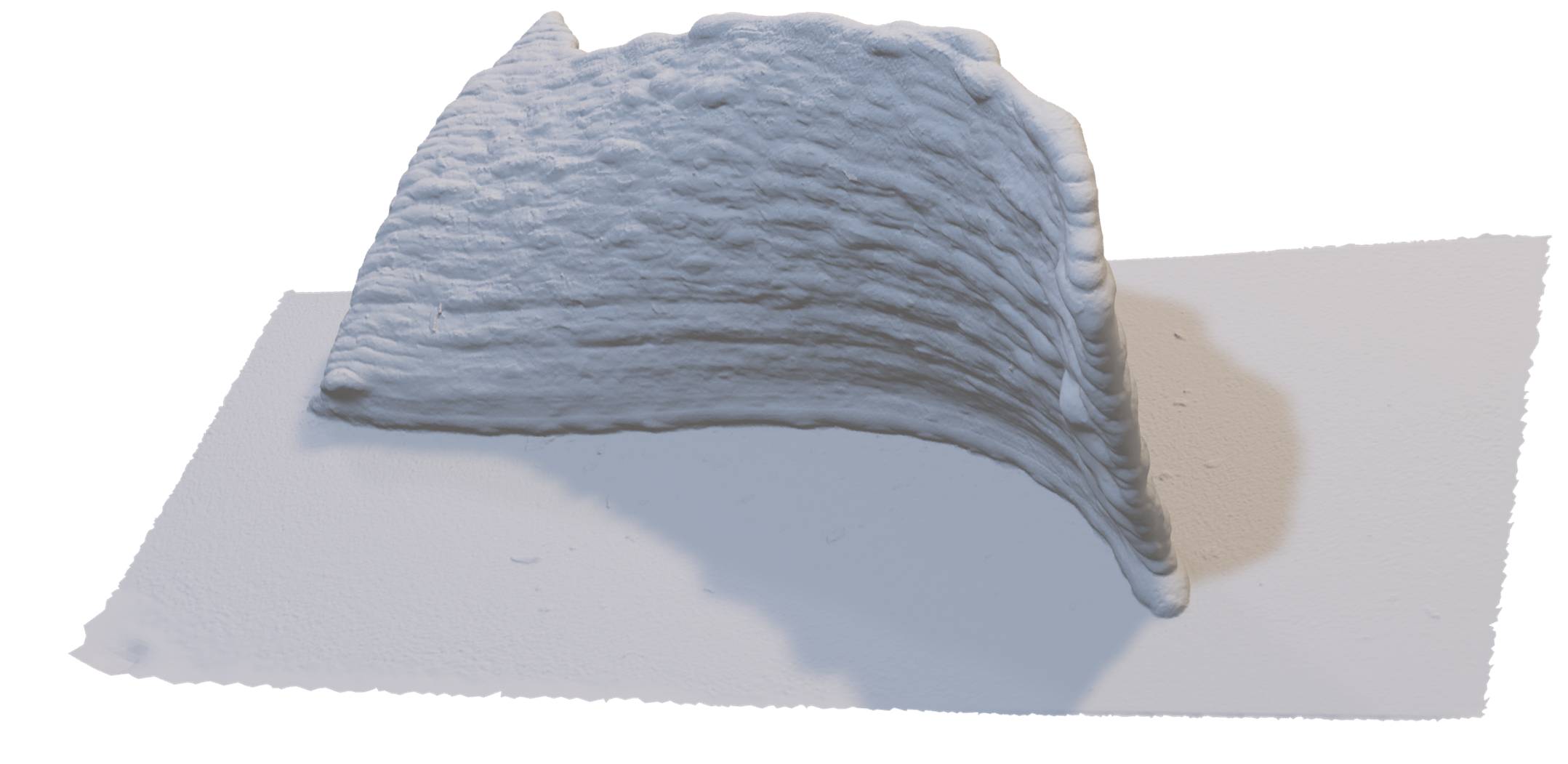}
     \caption{Scan Output Mesh}
    \end{subfigure}
    \begin{subfigure}[b]{0.5\textwidth}
     \centering
     \includegraphics[height=1.5in]{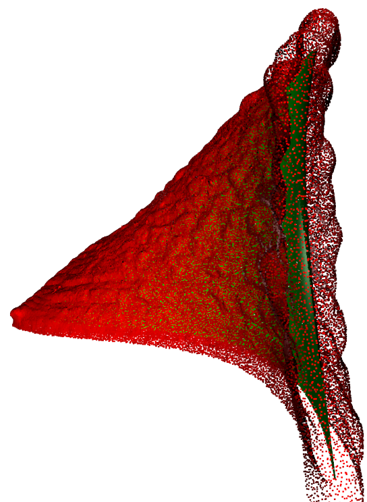}
     \caption{Aligned Pointcloud}
    \end{subfigure}
    \begin{subfigure}[b]{0.33\textwidth}
     \centering
     \includegraphics[height=1.5in]{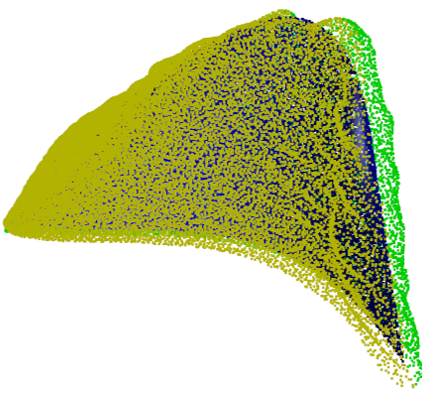}
     \caption{Processed Pointcloud}
     \label{fig:processed_pc}
    \end{subfigure}
    \caption{Product Evaluation with Artec Spider 3D Scanner.}
    \label{fig:artec_scan}
\end{figure}

After the completion of part printing using the robotic WAAM process, the resulting geometry may contain seams and defects due to inconsistent layer temperature, oxidation and other environmental effects. To quantitatively evaluate the WAAM performance, we use an Artec Spider 3D Scanner \cite{artec} to generate a 3D mesh file and compare it with the original CAD file as shown in Fig.~\ref{fig:artec_scan}.
Since this paper focuses on single-bead geometries, we only evaluate the overall thickness of the WAAM part. The scanned mesh is first converted to a point cloud and aligned with the original CAD using Iterative Closest Point (ICP) registration \cite{icp}.  The point cloud is then separated into left and right (inner and outer) edges. The width at each point of the surface is calculated by the perpendicular sum distance to the left and right edges as shown in Fig.~\ref{fig:processed_pc}.

\begin{table}[h!]
\centering
\begin{tabular}{||c c c c c||} 
 \hline
 Geometry & Material & $e_{avg}$ & $e_{max}$ & $\frac{\sigma(w)}{\mu(w)}$\\ [0.2ex] 
 & & mm & mm & \%  \\
 \hline\hline
 Bell & ER4043 (Aluminum) & 0.39 & 1.22 & 8.73 \\ 
 & ER70S-6 (Steel Alloy) & 0.48 & 1.60 & 13.45 \\   
 & ER316L (Stainless Steel) & 0.67 & 1.46 & 7.75 \\ 
 \hline
 Blade & ER4043 (Aluminum) & 0.25 & 1.69 & 11.22 \\ 
 & ER70S-6 (Steel Alloy) & 0.42 & 2.24 & 13.20 \\   
 & ER316L (Stainless Steel) & 0.25 & 1.65 & 12.57 \\ 
 \hline
\end{tabular}
\caption{WAAM Part Evaluation using Artec Spider 3D Scanner.}
\label{table:evaluation}
\end{table}

We use the following criteria to assess the shape accuracy of the printed part:
\begin{itemize}
    \item  $e_{avg}$: Average Euclidean distance error between the WAAM part and the original CAD 
    \item  $e_{\max}$: Maximum Euclidean distance error between the WAAM part and the original CAD 
    \item Width variation ($\frac{\sigma(w)}{\mu(w)}$\%): Standard deviation of the WAAM part bead width $\sigma(w)$ over the average of the WAAM part bead width $\mu(w)$.
\end{itemize}
The assessment based on these two criteria for the blade and cup geometries and all three materials are summarized in Table.~\ref{table:evaluation}.
%

All materials for both shapes are able to achieve below 3~mm worst-case error.  The source of shape formation errors includes robot kinematics, relative pose calibration, and possibly weld pool flow.  The width variance of the part is mostly due to the welding process such as arc striking/extinguishing, inconsistent standoff wire length, or inconsistent temperature of the part. Stainless steel tends to have better performance in both shape formation and width consistency, while aluminum has a larger worst-case error in shape formation.  


\section{Conclusion and Future Directions}
\label{sec:conclusion}
This paper presents the architecture, algorithms, and implementation results of an automated coordinated robotic WAAM process for single-bead complex geometries.  The demonstration testbed is based on Yaskawa Motoman robots and Fronius weld controller but the software architecture is applicable to other vendor systems.  The demonstration shows accurate printing of a variety of geometries, including generic turbine blade, bell, cup, funnel, and diamond, and materials, including aluminum, steel alloy, and stainless steel.  The overall art-to-part system includes single-bead geometry slicing, multi-robot coordinated trajectory planning, multi-sensor data streaming, and 3D product evaluation.  The system allows ready weld parameter adjustments including run-time torch speed and wire feed rate based on sensor data.
Our current work focuses on run-time motion correction for temperature and weld pool control based on sensor data, varying torch speed for non-uniform layer height, multi-bead printing, and active cooling.


\addtolength{\textheight}{-12mm}   




\section*{ACKNOWLEDGMENT}

Research was sponsored by the ARM (Advanced Robotics for Manufacturing) Institute through a grant from the Office of the Secretary of Defense and was accomplished under Agreement Number W911NF-17-3-0004. The views and conclusions contained in this document are those of the authors and should not be interpreted as representing the official policies, either expressed or implied, of the Office of the Secretary of Defense or the U.S. Government. The U.S. Government is authorized to reproduce and distribute reprints for Government purposes notwithstanding any copyright notation herein.


The authors would also like to thank Roger Christian, Chris Anderson and Ted Miller from Yaskawa Motoman, Pinghai Yang and Jeff Schoonover from GE Global Research, and Matt Robinson from Southwest Research Institute for their technical assistance and helpful discussion of the project.

\bibliographystyle{elsarticle-num}
\bibliography{bib}

\end{document}